%% file: main.tex
\documentclass{article}
\usepackage[T1]{fontenc}
\usepackage[utf8]{inputenc}
\usepackage{textcomp}
\usepackage{upquote}

\usepackage{listings}
\usepackage{xcolor}

\usepackage{microtype}
\usepackage{graphicx}
\usepackage{subcaption}
\usepackage{booktabs} 

\usepackage{hyperref}
\usepackage{enumitem}
\usepackage{xcolor} 
\usepackage{color}

\usepackage{colortbl}

\usepackage{pifont}

\usepackage[preprint]{icml2026}

\usepackage{amsmath}
\usepackage{amssymb}
\usepackage{mathtools}
\usepackage{amsthm}
\usepackage{wrapfig}
\usepackage{multirow}
\usepackage{listings}
\usepackage{xcolor}
\usepackage{pifont}
\usepackage{threeparttable}
\usepackage{listings}
\usepackage{xcolor}
\usepackage{tcolorbox}
\usepackage{longtable}
\usepackage{rotating} 
\usepackage{tikz}

\usepackage{tabularx}

\usepackage[capitalize,noabbrev]{cleveref}

\theoremstyle{plain}

\theoremstyle{definition}

\theoremstyle{remark}

\usepackage[textsize=tiny]{todonotes}

\icmltitlerunning{CATArena: Evaluating Evolutionary Capabilities of Code Agents via Iterative Tournaments}

\begin{document}

\twocolumn[
  \icmltitle{CATArena: Evaluating Evolutionary Capabilities of Code Agents via Iterative Tournaments}



  \icmlsetsymbol{equal}{*}

  \begin{icmlauthorlist}
    \icmlauthor{Lingyue Fu}{equal,sjtu}
    \icmlauthor{Xin Ding}{equal,sjtu}
    \icmlauthor{Linyue Pan}{thu}
    \icmlauthor{Yaoming Zhu}{meituan}
    \icmlauthor{Shao Zhang}{sjtu}
    \icmlauthor{Lin Qiu}{meituan}
    \icmlauthor{Weiwen Liu}{sjtu}
    \icmlauthor{Weinan Zhang}{sjtu}
    \icmlauthor{Xuezhi Cao}{meituan}
    \icmlauthor{Xunliang Cai}{meituan}
    \icmlauthor{Jiaxin Ding}{sjtu}
    \icmlauthor{Yong Yu}{sjtu}
  \end{icmlauthorlist}

 \icmlaffiliation{sjtu}{Shanghai Jiao Tong University, Shanghai, China}
 \icmlaffiliation{meituan}{Meituan, China}
 \icmlaffiliation{thu}{Shenzhen International Graduate School, Tsinghua University, Shenzhen, China}

  \icmlcorrespondingauthor{Weiwen Liu}{wwliu@sjtu.edu.cn}
  \icmlcorrespondingauthor{Lingyue Fu}{fulingyue@sjtu.edu.cn}
  
  \icmlkeywords{Code Agents, Evaluation, Evolutionary Capabilities}

  \vskip 0.3in
]



\printAffiliationsAndNotice{}  

\input{section/Abstract}

\input{section/new_intro}

\input{section/Related}

\input{section/Benchmark}

\input{section/Experiment}
\input{section/Conclusion}
\clearpage
\newpage
\section*{Impact Statement}

This paper presents CATArena, a benchmark for evaluating the evolutionary capabilities of code agents. Our goal is to advance the reliability and efficiency of automated software engineering. There are many potential societal consequences of our work, none of which we feel must be specifically highlighted here.

\bibliography{ref}
\bibliographystyle{icml2026}

\newpage
\appendix
\onecolumn

\input{section/Appendix}




\end{document}

%% file: section/Abstract.tex
\begin{abstract}
Current evaluation for Large Language Model (LLM) code agents predominantly focus on generating functional code in single-turn scenarios, which fails to evaluate the agent's capability for continuous code optimization and multi-turn iterative development. To bridge this gap, we introduce CATArena, a framework designed to evaluate the evolutionary capabilities of code agents via iterative tournaments. Agents engage in multi-turn tournaments and continuously refine their code through self-reflection and peer-learning based on comprehensive execution feedback. For evaluation, we propose a dual-metric system to decouple static generation proficiency from evolutionary potential. Extensive experiments reveal that an agent's evolutionary potential is not strictly correlated with its initial proficiency. Our analysis further reveals that current agents struggle to concurrently leverage both peer-learning and self-reflection for effective performance gains. Furthermore, the results validate CATArena's high extensibility and resistance to variance tasks, establishing it as a continuous and reliable standard for assessing the evolutionary capability of LLM code agents.~\footnote{The official website of CATArena is \url{https://catarena.ai}.}
\end{abstract}

%% file: section/new_intro.tex
\section{Introduction}
The rapid evolution of Large Language Models (LLMs) has catalyzed the emergence of code agents. Unlike passive chat assistants, code agents actively interact with dynamic digital workspaces, manipulating file systems and utilizing toolchains to orchestrate complex software development workflows~\cite{dong2025surveycodegenerationllmbased,dearing2025leveragingllmsautomateenergyaware,xu2025hallucinationconsensusmultiagentllms}. To assess these expanding capabilities, the evaluation landscape has evolved from simple method-level generation~\cite{chen2021evaluatinglargelanguagemodels} to repository-level engineering scenarios~\cite{xu2025swecompassunifiedevaluationagentic,xu2025webbenchllmcodebenchmark}. Performance is predominantly quantified by static metrics such as task success rate~\cite{yang2024sweagentagentcomputerinterfacesenable} within a single-turn development process. While some emerging benchmarks incorporate non-functional metrics such as security or code complexity, these remain static checks rather than dynamic objectives.

These static and correctness-centric benchmarks impose two critical limitations on advancing agent intelligence. \textbf{First and foremost, existing protocols suffer from the absence of iterative evolution.} Real-world software development is inherently incremental and collaborative~\cite{BENNETT1996673}: human engineers continuously refine their solutions by analyzing execution logs and learning from peer reviews~\cite{10.1145/3613905.3651008}. Crucially, this evolutionary mechanism is equally indispensable for autonomous agents. However, by restricting evaluation to a single-turn mode, current frameworks isolate agents from the dynamic feedback loops essential for optimization and evolution. Agents are prevented from rectifying runtime errors through self-debugging or optimizing strategies by observing superior peer implementations~\cite{gao2025survey,zhu2025evolutionaryperspectivesevaluationllmbased}. Consequently, current benchmarks capture only a static snapshot of the agent's baseline competency, failing to evaluate its dynamic capacity for self-reflection and peer-learning over time.

Furthermore, \textbf{this lack of iterative depth leads to an underemphasis on code quality optimization.} Since evaluation is confined to a single turn, current frameworks predominantly rely on binary functional correctness metrics like pass@k~\cite{chen2021evaluatinglargelanguagemodels,austin2021programsynthesislargelanguage}, failing to incentivize agents to refine code for computational efficiency or robustness. In large-scale software engineering, a functionally correct but inefficient algorithm can lead to unacceptable system latency or energy consumption~\cite{dearing2025leveragingllmsautomateenergyaware}. However, standard benchmarks treat a brute-force solution as equivalent to an optimized expert implementation. This equivalence masks the performance gap between agents that merely solve a problem and those that engineer superior solutions, causing metrics to saturate without capturing the potential for unbounded evaluation.

To bridge this gap, we introduce CATArena (\textbf{C}ode \textbf{A}gent \textbf{T}ournament \textbf{Arena}), a dynamic evaluation framework designed to assess the evolutionary capabilities of code agents through iterative refinement. Unlike traditional static benchmarks, CATArena immerses agents in a multi-turn development loop, mimicking the continuous improvement process of human engineers. Within this interactive workspace, agents must go beyond generating functional code. They are required to analyze execution feedback and, crucially, leverage peer solutions and opponent strategies to iteratively optimize their performance. To ensure an unbounded evaluation, we instantiate six diverse tasks with high performance ceilings and design scoring mechanisms that reward continuous optimization. These tasks are categorized into two types based on the feedback signal: two \textit{objective optimization tasks}, where agents optimize against fixed metrics (e.g., time complexity, throughput); and four \textit{competitive tasks}, a novel introduction where agents adaptively optimize their strategies against dynamic opponents (e.g., in strategy games, network attack and defense). This dual-track design allows the framework to seamlessly integrate diverse task types while effectively measuring an agent's potential to evolve from basic correctness to expert-level excellence.

We conduct extensive experiments using a multi-round iterative process, evaluating both self-developed minimal agents powered by various LLMs and state-of-the-art (SoTA) commercial code agents. Our analysis reveals that evolutionary capability is a distinct dimension of intelligence, independent of single-turn generation capability. We observe that agents with high initial pass rates do not necessarily possess strong evolving abilities. Conversely, some agents demonstrate low-start, high-growth trajectories, effectively leveraging feedback to surpass initially superior competitors. Furthermore, by analyzing the optimization trajectories, we identify distinct evolutionary patterns in how agents utilize peer solutions and execution logs. Ultimately, these results demonstrate that the iterative framework enables agents to overcome the performance plateaus typical of single-round development, validating the necessity of unbounded evaluation for the next generation of intelligent agents.

In summary, our contributions are as follows:
\begin{itemize}[leftmargin=10pt]
    \item \textbf{Evolutionary Capability Evaluation}:  To the best of our knowledge, we are the first to identify and formalize the evolutionary capability of code agents, aligning with the iterative nature of real-world engineering.
   
    \item \textbf{CATArena Framework}: We introduce CATArena, a tournament-based evaluation framework that models the development process as a multi-turn evolutionary loop. Driven by iterative feedback, it quantifies code refinement via self-reflection and peer-learning, transcending static, single-turn evaluation.

    \item \textbf{Comprehensive Agent Analysis}: We demonstrate that evolutionary capability is orthogonal to static generation performance and analyze its temporal dynamics across tasks. Our extensible experiments further highlights CATArena’s extensibility with new environments, revealing that SoTA agents still faces significant challenges on complex tasks.
\end{itemize}

%% file: section/Related.tex
\section{Related Work}
\begin{figure*}[t]
    \centering
    \includegraphics[width=\linewidth]{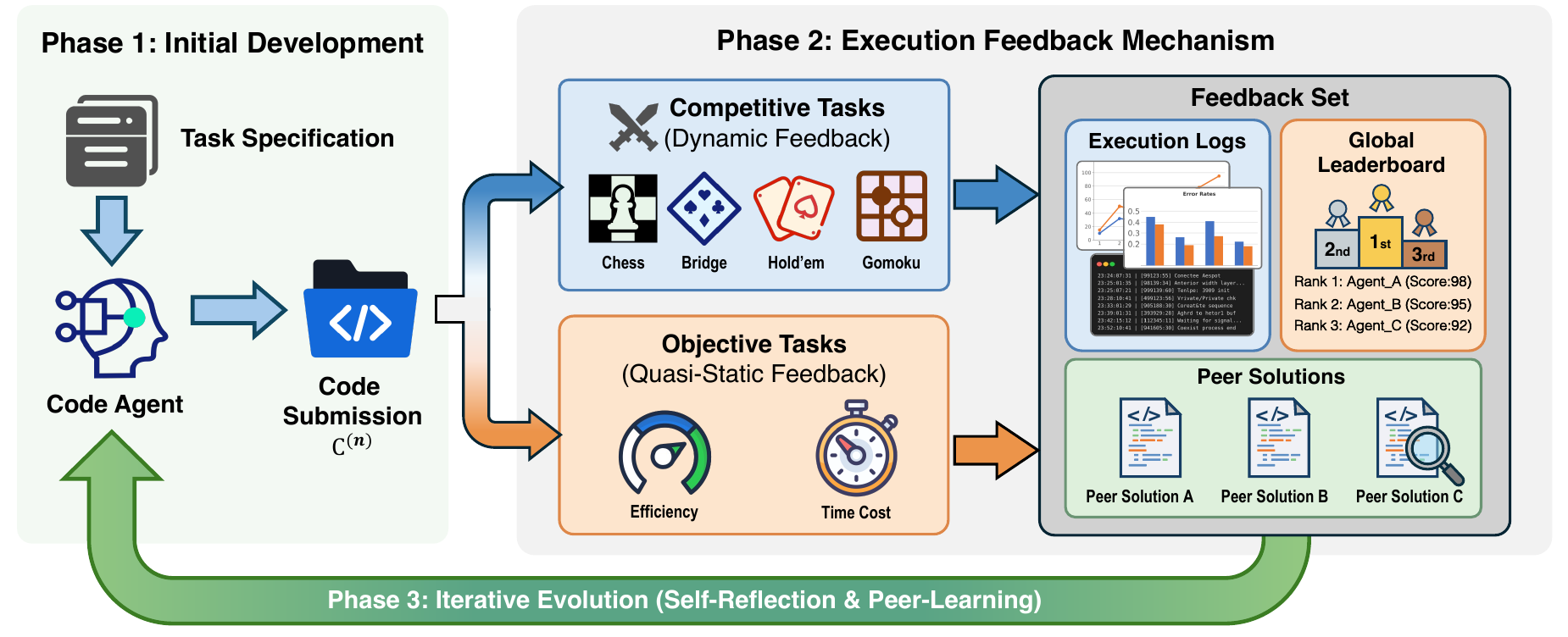}
    \caption{\textbf{Overview of the CATArena Framework.} (1) \textbf{Initial Development}: Agents receive task specifications and generate initial solution code $C^{(1)}$. (2) \textbf{Execution Feedback}: Submissions are evaluated in either competitive or objective tracks. Results are aggregated into a structured set containing execution logs, leaderboards, and peer solutions. (3) \textbf{Iterative Evolution}: Agents engage in self-reflection and peer-learning to refine their code $C^{(n)}$ for the subsequent rounds. }
    \label{fig:overview}

\end{figure*}

\textbf{Evolutionary Mechanisms.}
Evolutionary capability is pivotal for agents to adapt in dynamic environments. For LLM-based agents, this manifests as in-context evolution, where agents optimize strategies using historical context and execution feedback without parameter updates~\cite{zhu2025evolutionaryperspectivesevaluationllmbased}. Recent research highlights two primary mechanisms driving this evolution: self-reflection and peer-learning. Self-reflection methods~\cite{madaan2023self,shinn2023reflexion} enhance outputs through iterative self-correction based on environmental feedback~\cite{you2024llm}. Peer-learning approaches encourage agents to distill insights from the reasoning processes and solutions of others~\cite{liang-etal-2024-encouraging,luo2025learning}. While these mechanisms have advanced code generation and reasoning, the systematic evaluation of an agent's evolutionary capability remains underexplored, particularly regarding how agents leverage feedback to evolve over time in competitive coding scenarios. CATArena fills this void as the first to assess the evolutionary capability of code agents.

\textbf{Evaluation of Code Agents.}
Recent benchmarks have scaled evaluation to simulate complex, real-world software engineering scenarios. In the domain of repository-level development, frameworks such as GitTaskBench~\cite{GitTaskBench}, SWE-PolyBench~\cite{SWE-PolyBench}, and SWT-Bench~\cite{swtbench} challenge agents with large-scale maintenance and bug fixing, complementing established standards like SWE-bench~\cite{yang2024sweagentagentcomputerinterfacesenable}. Beyond general engineering, specialized evaluations have emerged for diverse domains: RedCode~\cite{RedCode} and SecBench~\cite{lee2025secbenchautomatedbenchmarkingllm} focus on code security; InfiAgent-DABench~\cite{InfiAgent-DABench} and DA-Code~\cite{huang2024code} target data science workflows; while other works assess research automation~\cite{DBLP:journals/corr/abs-2506-11763} and web development~\cite{xu2025webbenchllmcodebenchmark}. Despite this extensive coverage, the evaluation methodology remains predominantly static. Performance is typically quantified by binary success metrics such as Pass@k~\cite{chen2021evaluatinglargelanguagemodels} and task success rate~\cite{yang2024sweagentagentcomputerinterfacesenable}. Although some benchmarks incorporate non-functional metrics like code complexity~\cite{10.1007/978-3-032-12089-2_26} or test coverage~\cite{chen2025evaluating}, these are applied as static checks. 

To overcome the limitations of static metrics, research has expanded into open-ended tasks. In the realm of strategic reasoning, benchmarks like GameBench~\cite{GameBench} and TextArena~\cite{guertler2025textarena} assess agents through multi-turn gameplay. Similarly, for code optimization, datasets such as OIBench~\cite{zhu2025oibenchbenchmarkingstrongreasoning} and SWE-bench~\cite{yang2024sweagentagentcomputerinterfacesenable} provide open-ended challenges for algorithmic reasoning and engineering performance. However, current evaluations for these tasks are predominantly single-turn, treating them as static problems rather than opportunities for continuous optimization. This restriction prevents code agents from demonstrating their full potential through iterative refinement. CATArena bridges this gap by incorporating these open-ended tasks into an evolutionary framework, rigorously testing the limits of code agents.

%% file: section/Benchmark.tex
\section{CATArena Framework}
In this section, we detail the CATArena framework, a universal and robust evaluation paradigm designed to shift the assessment of code agents from static correctness to dynamic evolutionary capability. 

\subsection{Iterative Evolution via Tournament Feedback}
\label{subsec:framework}

Unlike traditional benchmarks that treat coding as a static, single-turn task, CATArena abstracts the evaluation process into a continuous loop of development, execution, and evolution. As illustrated in Figure~\ref{fig:overview}, this paradigm is structured into two primary phases bridged by an execution feedback mechanism.

\input{tables/supportedGames}
\textbf{Initial Development (Round 1).}
The process begins with the initial development phase, which establishes the single-turn development capability of the code agent. In this stage, the agent receives a task specification $\mathcal{T}$, ranging from game rules to algorithmic problems, along with a minimal functional example. Operating without external guidance or historical data, the agent must comprehend the requirements and implement an initial solution $ C^{(1)}$. 

\textbf{Execution Feedback.}
Upon receiving the code submission, CATArena initiates the execution feedback process to drive the transition from static code generation and dynamic evolution. The system executes the submitted code within a sandboxed evaluation arena to generate quantitative performance scores. This phase aggregates execution results into a structured feedback set containing comprehensive tournament execution logs, a global leaderboard, and peer solutions. While the specific metrics and feedback modalities vary depending on the task type (as detailed in Section~\ref{subsec:tasks}), the fundamental function remains consistent in providing the necessary signals for evolution.

\textbf{Iterative Evolution (Round $n > 1$).}
In the iterative evolution phase, the goal of agents shifts from generation to optimization. Agents are provided with the task specification and the execution feedback derived from the previous round. By analyzing this information, agents are tasked to identify weaknesses in their previous implementations via self-reflection and assimilate superior approaches through peer-learning. The agent then synthesizes these insights to generate an evolved solution $C^{(n)}$. This cycle repeats for $N$ rounds, simulating the real-world engineering process where solutions are continuously refined based on testing and analysis of external implementations.


\subsection{Task Categorization and Instantiation}
\label{subsec:tasks}

To comprehensively evaluate the evolutionary potential of code agents, CATArena instantiates six diverse coding tasks. As demonstrated in Table~\ref{tab:tasks}, these tasks are categorized into two distinct types based on the nature of the feedback source, mirroring the duality of real-world software engineering where development is driven either by adversarial dynamics or distinct system efficiency targets: \textit{Competitive Tasks} and \textit{Objective Optimization Tasks}.

\textbf{Competitive Tasks.}
This category models scenarios where performance is inherently relative. Unlike static benchmarks, the feedback mechanism here is dynamic: a solution's effectiveness depends entirely on the strength of its opponents. As the population evolves, the difficulty of winning increases, creating a non-stationary evaluation target that compels agents to continuously adapt to new counter-strategies. To capture this, we employ a Round-Robin tournament structure where the code implementations submitted by agents are executed against each other in pairwise or multi-agent matches. Crucially, the feedback provided to agents is not limited to their own match results. It encompasses the full tournament logs and peer solutions from the entire population. We instantiate four specific arenas: Gomoku, Texas Hold'em, Chess, and Bridge. This diversity, covering both symmetric perfect-information games and asymmetric imperfect-information settings, rigorously tests the agent's ability to analyze complex interaction dynamics and refine strategies against an adaptive field.

\textbf{Objective Optimization Tasks.}
This category models scenarios where the primary goal is to optimize a static objective function. Here, the feedback is quasi-static. While the evaluation metrics are deterministic, the competitive standard is continuously raised by the population. Agents must treat functional correctness merely as a starting point, striving to approach theoretical limits to outperform peer solutions. We incorporate two specific benchmarks. First, \textit{OIBench}~\cite{zhu2025oibenchbenchmarkingstrongreasoning} evaluates algorithmic efficiency using Time AUC, a metric that quantifies the volume of test cases passed within constrained time limits. Second, \textit{SWE-Perf}~\cite{he2025sweperflanguagemodelsoptimize} assesses repository-level engineering by challenging agents to optimize authentic GitHub codebases. It measures the performance speedup relative to the original implementation, highlighting the gap between functional code and expert-level optimization.

\subsection{Formalization and Metrics}
\label{subsec:metrics}

To provide a standard of unified evaluation across diverse tasks, we construct a global performance matrix $W$. Based on this matrix, we derive two core metrics to comprehensively assess both the agent's static coding capability and its evolutionary potential.

\textbf{Performance Matrix Construction.}
Let $N$ be the total number of rounds and $M$ be the number of participating agents. We define the performance matrix $W \in \mathbb{R}^{(M \times N) \times (M \times N)}$, where $W_{i,j}^{n,m}$ denotes the performance score of Agent $i$ (from Round $n$) evaluated against Agent $j$ (from Round $m$).  

\begin{itemize}[leftmargin=10pt]
    \item For competitive tasks, $W_{i,j}^{n,m}$ represents the win rate of agent $i$'s code in round $n$ against agent $j$'s code in round $m$. For symmetric settings, this is derived from pairwise round-robin matches. For asymmetric settings (e.g., Texas Hold'em), we utilize batch-based tournaments where scores reflect the win probability in the tournament. 
    \item For objective optimization tasks, since agents operate independently. Thus, we simplify the notation $W_{i,j}^{n,m}$ to $W_i^n$, representing the scalar performance score of agent $i$ at iteration $n$. Specifically, for \textit{OIBench}, we adopt their Time AUC metric~\cite{zhu2025oibenchbenchmarkingstrongreasoning}, which quantifies the area under the curve of passed test cases against varying time constraints, normalized by the maximum achievable area.  For \textit{SWE-Perf}, we utilize the performance ratio~\cite{he2025sweperflanguagemodelsoptimize}, defined as the computational speedup of the optimized code relative to the original baseline, normalized by a target expert reference.
\end{itemize}
Detailed formalization of the matrix $W$ in CATArena is provided in Appendix~\ref{app:W_calc}. To mitigate randomness in game outcomes, every entry in the scoring matrix $W$ is estimated by repeated matches for fixed code submissions.  Detailed repetition settings for each task are provided in Appendix~\ref{app:tournamentFormat}.

\textbf{Basic Performance $S_{\text{base}}$.}
This metric evaluates the agent's static coding ability in a traditional single-turn setting. It quantifies the relative performance of the agent's initial submission $C^{(1)}$ within the Round 1 population:
\begin{equation}
    S_{\text{base}, i} = \frac{1}{M-1} \sum_{j \neq i} W_{i,j}^{1,1}.
\end{equation}
For objective tasks, this simplifies to the score of the first round $W_{i}^{1}$. $S_{\text{base}}$ serves as a baseline to reflect the agent's zero-shot task-solving proficiency before any feedback is introduced.

\paragraph{Evolutionary Capability $S_{\text{evo}}$.}
This metric is the primary indicator of evolutionary capability. Unlike static metrics, it quantifies the rate of improvement driven by feedback. 

We first calculate the global performance score $G_{i}^n$ for each round $n$. For competitive tasks, $G_{i}^n$ represents the relative strength of Agent $i$'s Round $n$ code when evaluated against the entire population of code versions ($M \times N$) produced across all rounds:
\begin{equation}
    G_{i}^n = \frac{1}{MN-1} \sum_{(j,m) \neq (i,n)} W_{i,j}^{n,m}.
\end{equation}
We use global average score for $G_{i}^n$ and give a brief explanation in Appendix~\ref{app:elo_vs_g} about the stability of this metric.
For objective tasks, $G_{i}^n$ is simply the score $W_{i}^{n}$. 

$S_{\text{evo}, i}$ is derived as the slope $k_i$ of the linear fit $G_i^n \approx k_i \cdot n + b_i$ using the Ordinary Least Squares (OLS) method:
\begin{equation}
    S_{\text{evo}, i} = \frac{\sum_{n=1}^{N} (n - \bar{n})(G_i^n - \bar{G}_i)}{\sum_{n=1}^{N} (n - \bar{n})^2},
\end{equation}
where $\bar{n}$ and $\bar{G}_i$ are the average round number and average global performance, respectively.
A positive $S_{\text{evo}}$ indicates effective adaptation, with higher values signifying a faster rate of evolution.

We employ OLS method primarily because development failures might occur in any round due to inherent limitations in their robustness. Compared to simpler alternatives such as total improvement (i.e., $G^N - G^1$), which are highly sensitive to single-round outliers, OLS provides a more robust estimation of the overall growth trend by mitigating the impact of transient failures. Importantly, we regard development failures as a failure of the evolutionary process itself, and these instances are explicitly factored into the OLS calculation to reflect the agent's inability to maintain stable progress. Furthermore, as our experiments focus on the initial active evolution phase ($N \leq 4$), a linear approximation effectively captures the rate of evolution before performance saturation.

%% file: tables/supportedGames.tex
\begin{table*}[t]
\centering
\small
\caption{\textbf{Overview of tasks in CATArena.} The framework instantiates six diverse tasks. Crucially, the feedback mechanism provides agents with population-level data, including full tournament logs and peer solutions, rather than just individual execution results, fostering group evolution.}
\label{tab:tasks}
\resizebox{0.95\textwidth}{!}{
\begin{tabular}{l|l l c c}
\toprule
\textbf{Category} & \textbf{Task} & \textbf{Domain} & \textbf{Players} & \textbf{Feedback Content} \\
\midrule
\multirow{4}{*}{\textbf{Competitive}} 
& Gomoku & Board Game & 2 & \multirow{4}{*}{Adversarial Match Logs \& Peer Solutions} \\
& Texas Hold'em & Card Game & $\geq 8$ & \\
& Chess & Board Game & 2 & \\
& Bridge & Card Game & 4 & \\
\midrule
\multirow{2}{*}{\textbf{Objective}} 
& OIBench~\cite{zhu2025oibenchbenchmarkingstrongreasoning} & Algorithm Optimization & 1 & \multirow{2}{*}{Deterministic Efficiency Metrics \& Peer Solutions} \\
& SWE-Perf~\cite{he2025sweperflanguagemodelsoptimize} & Software Optimization & 1 & \\
\bottomrule
\end{tabular}
}
\end{table*}

%% file: section/Experiment.tex
\section{Experiments}
\label{sec:experiments}
In this section, we conduct comprehensive experiments to validate the effectiveness of the CATArena framework and analyze the evolutionary behaviors of code agents under the proposed iterative paradigm. Specifically, our evaluation aims to address the following four research questions:
\begin{itemize}[leftmargin=10pt]
    \vspace{-5pt}
    \item \textbf{RQ1:}  Can CATArena separate evolutionary capability ($S_{evo}$) from static base performance ($S_{base}$)?
    
    \item \textbf{RQ2: } What dynamics do agents exhibit over iterations?
    
    \item \textbf{RQ3:} How do peer-learning and self-reflection affect evolution?
    
    \item \textbf{RQ4:} Is CATArena robust to task variants and extensible to new tracks?
    \vspace{-5pt}
\end{itemize}

\input{tables/RQ1}

\subsection{Experimental Setups}
\label{sec:experimental_setup}

\paragraph{Participants.} We evaluate two categories of agents to disentangle model reasoning from system engineering: (1) \textbf{Minimal Agents:} To isolate the intrinsic evolutionary potential of foundation models, we implement a standardized minimal agent framework using Google Agent Development Kit (ADK)\footnote{https://github.com/google/adk-python}. This minimalist setup provides only core tools (e.g., file I/O and scripting), ensuring that performance differences are primarily attributable to the underlying LLM rather than complex scaffolding~\cite{kim2023language,shihipar2026claude}. We select a diverse set of API-based and open-weight models as backbones. (2) \textbf{Commercial Code Agents:} We include leading CLI-based commercial agents to establish high-performance industry baselines. These systems feature proprietary toolchains and represent the current state-of-the-art in agentic development. For brevity, we refer to agents by their model family names; full specifications are provided in Appendix~\ref{app:generationconfig}.

\paragraph{Evaluation Protocol.} All experiments are conducted within two distinct cohorts to facilitate intra-group evaluation:
(1) A comparison among minimal agents equipped with various LLMs, and 
(2) A comparison between the best-performing minimal agent and a suite of commercial code agents. 
This setup ensures that agents are evaluated against peers within their respective categories to maintain a fair and reliable assessment of evolutionary capability.

\textbf{Repetition and Stability.} Due to the high computational and token costs of long-term evolution, the main evolutionary experiments are conducted as a single chain. To validate the reliability of this approach, we conducted a separate \textit{stability analysis} on the initial rounds (Rounds 1 and 2), repeating the code generation process four times. We observe that while the specific code implementations vary across runs, the relative performance rankings among agents remain highly consistent (see Appendix~\ref{sec:repeat_exp} for analysis).

\subsection{Results of Evolutionary Evaluation (RQ1)}
\label{sec:rq1}

\begin{figure}[t]
    \centering
   \includegraphics[width=\linewidth]{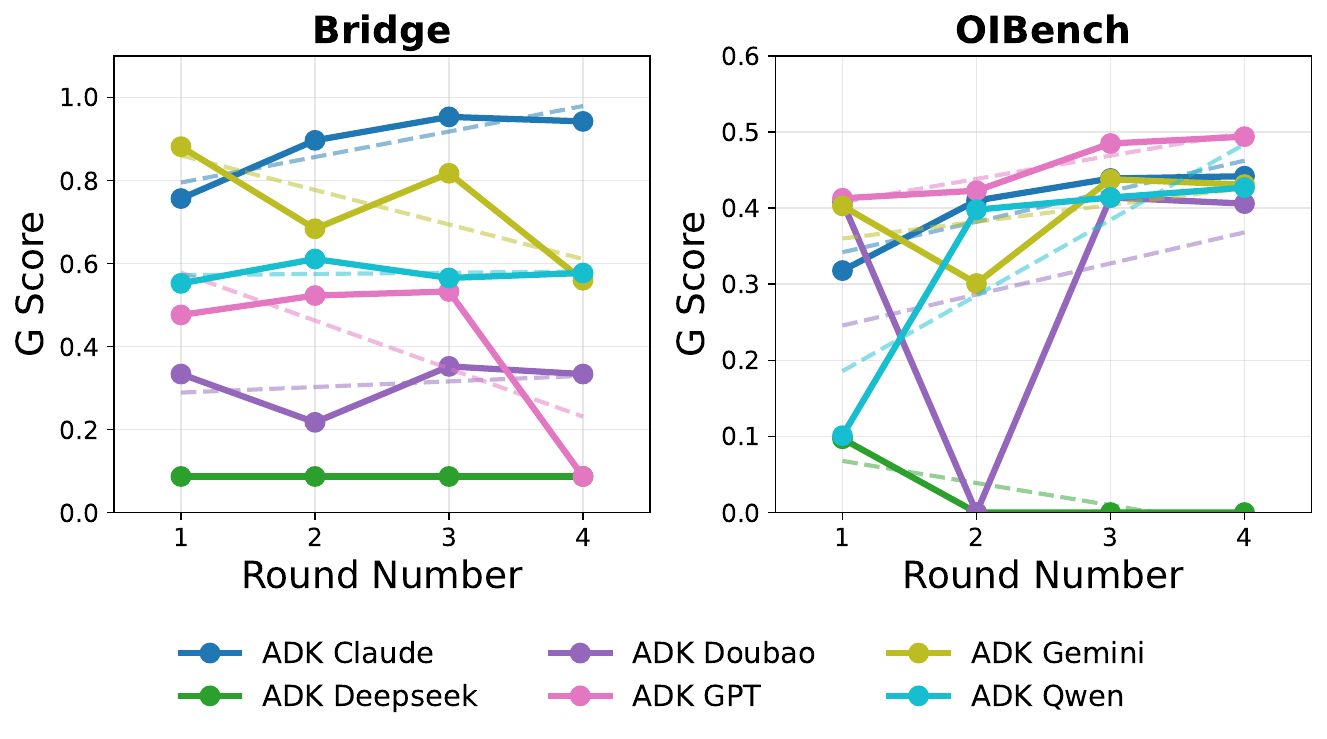}
    \caption{\textbf{Evolutionary Trajectories across Iterations.} Solid lines represent the actual $G$ Score, dashed lines indicate the linear regression fits used to calculate $S_\text{evo}$.}
    \vspace{-12pt}
    \label{fig:evolution_trajectory}
\end{figure}

\textbf{Overall Performance Comparison.}
Table~\ref{tab:rq1} summarizes the Base Performance ($S_\text{base}$) and Evolutionary Capability ($S_\text{evo}$) across all agents. We also report statistics on HTTP error rates in Appendix~\ref{app:httpError}. The results yield three key observations: (1) \textit{We observe a weak correlation between $S_\text{base}$ and $S_\text{evo}$ across all tasks.} High $S_\text{base}$ does not guarantee effective evolution. For instance, while Claude-4 dominates the initial round in Chess ($0.90$), it suffers significant regression ($-0.061$) during evolution. Conversely, Gemini-2.5-Pro demonstrates a low-start, high-growth trajectory in Gomoku ($0.25 \rightarrow +0.156$). While it starts with a deficit, its high $S_\text{evo}$ indicates the ability to rapidly absorb feedback.
(2) \textit{Evolutionary dynamics are inherently task-dependent.} We observe a distinct complexity barrier in the evolution of agents. In high-difficulty environments such as {Chess} and {SWE-Perf}, most agents exhibit negligible or negative $S_\text{evo}$. This indicates that when the reasoning gap between the current implementation and the optimal solution exceeds a certain threshold, agents struggle to extract actionable insights from feedback.
(3) \textit{Commercial frameworks primarily enhance static generation rather than evolution ability.} While commercial agents frequently outperform the Best ADK Agent in $S_\text{base}$, they do not exhibit a decisive advantage in $S_\text{evo}$. This disparity underscores that evolutionary capability is a distinct dimension of intelligence, largely unaddressed by current commercial scaffolding, and represents a critical frontier for future agentic system research.

\textbf{Evolutionary Trajectories.}
Figure~\ref{fig:evolution_trajectory} visualizes the round-wise performance dynamics of global performance $G^n$, highlighting three critical phenomena.
First, we observe significant \textit{performance oscillation} across iterations. This volatility indicates that current agents lack robust self-evaluation mechanisms, often discarding valid strategies for inferior ones during the evolutionary process. Second, despite this instability, evolution can drive \textit{substantial performance leaps}. For instance, in OIBench, {Qwen} (blue line) advances from a last-place initialization ($G_1= 0.10$) to the top-performing tier ($G_4 = 0.43$) through iterative refinement. This trajectory confirms that static zero-shot benchmarks may underestimate an agent's true potential, which can be significantly amplified through iterative feedback. Third, the visualization validates the reliability of our metric design. As indicated by the dashed regression lines, {$S_\text{evo}$ effectively captures the global optimization trend}, proving robust to transient performance drops caused by exploratory failures typical of stochastic code generation.
The results for other tasks are detailed in Appendix~\ref{app:learning}.

\subsection{Temporal Dynamics of Evolution (RQ2)}
\label{sec:rq2}
We investigate the temporal dynamics of agent evolution by analyzing population statistics across task types and extending the evolutionary horizon to $N=7$ rounds.

\textbf{Population Evolution Patterns.}
Table~\ref{tab:dynamics} characterizes the evolutionary trends using two metrics derived from Pearson correlation coefficients: $\text{Trend}_\text{mean}$, representing the correlation between round number and average population performance, and $\text{DIS}_\text{std}$, the correlation between round number and score dispersion (standard deviation). The results reveal that the population structure is heavily dictated by the feedback mechanism. Objective optimization tasks tend to drive \textit{divergence} ($\text{DIS}_\text{std} > 0$), whereas competitive tasks foster \textit{convergence} ($\text{DIS}_\text{std} < 0$). This disparity arises because objective feedback rewards strong agents capable of approaching theoretical limits, amplifying the gap, while the open-source nature of competitive tournaments allows weaker agents to mimic successful strategies, homogenizing the population. Regarding overall performance, the majority of tasks exhibit a positive $\text{Trend}_\text{mean}$, confirming that iterative refinement generally drives collective improvement. The only exception is Chess ($\text{Trend}_\textbf{mean} < 0$), where the excessive reasoning complexity overwhelms the feedback loop, leading to a collapse rather than evolution.

\input{tables/task}

\begin{figure}[t]
    \centering
    \includegraphics[width=\linewidth]{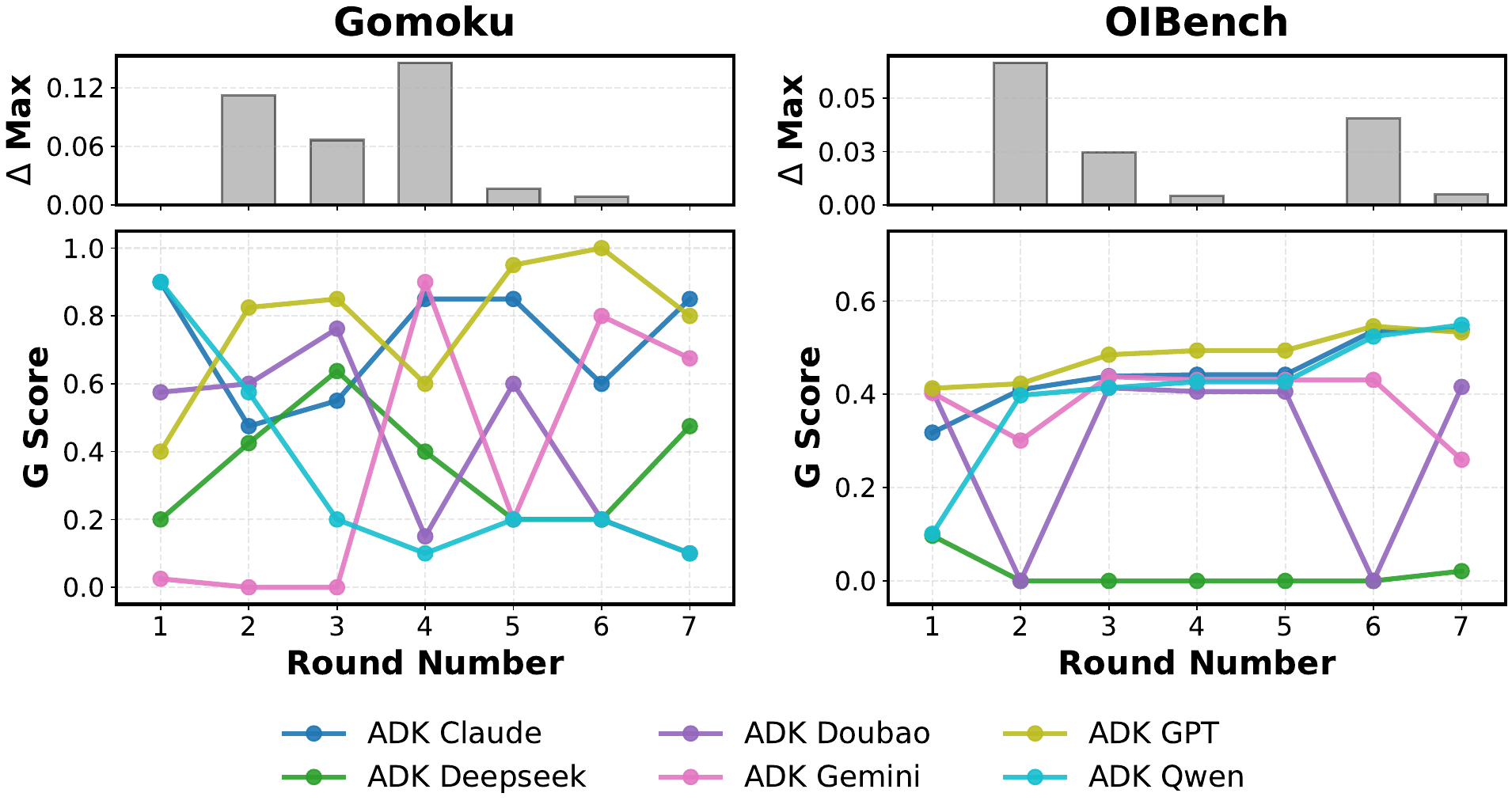}
    \caption{\textbf{Long-term Evolutionary Dynamics ($N=7$).} \textit{Top}: The Peak Performance Gain $\Delta \text{Max}$ across iterations. \textit{Bottom:} The trajectories of Global Performance $G^n$ for individual agents.}
    \label{fig:long_term_evolution}
    \vspace{-8pt}
\end{figure}

\textbf{Long-term Evolution.}
We extend the evaluation to $N=7$ for Gomoku and OIBench to investigate long-term evolution, visualizing the results in Figure~\ref{fig:long_term_evolution}. We first analyze the Peak Performance Gain ($\Delta \text{MAX}$), defined as the sum of performance increments across all models between consecutive rounds, which quantifies the evolutionary vitality. We observe that $\Delta \text{MAX}$ is significantly higher during the initial rounds ($N < 4$) compared to the later stages ($N > 4$). This decay justifies our selection of a four-round window for the main experiment, as it effectively captures the most active phase of evolution. Furthermore, the global performance $G^n$ trajectories in the bottom panel reveal a clear pattern of stratification. While early rounds exhibit frequent rank crossings, the rankings stabilize in the long term with significantly fewer intersections. Most agents eventually trend towards saturation, failing to break through performance ceilings in the later rounds. This suggests that while iterative feedback optimizes strategies, the asymptotic performance remains bounded by the intrinsic reasoning limits of the underlying models.

\subsection{Mechanism of Evolution (RQ3)}
\label{sec:rq3}
To deconstruct the black box of agent evolution, we categorize the evolutionary process into two distinct mechanisms: \textbf{Self-Reflection}, where the agent introspectively analyzes its own execution logs to fix bugs and refine logic; and \textbf{Peer-Learning}, where the agent extrospectively analyzes external solutions to adopt superior strategies or algorithms.

\input{tables/ablation_PL}
\textbf{Ablation Study.}
To isolate the contribution of two evolution mechanisms, we compare agent performance  with full feedback ($G_2$) against a self-reflection-only baseline ($G_2'$). We omit the symmetric ablation (peer-only without logs) as it reduces agents to blind imitation lacking performance grounding. As detailed in Table~\ref{tab:ablation_PL}, the results reveal a critical limitation: \textit{current agents struggle to synergize peer code and execution logs.} Instead of benefiting from a unified feedback loop, agents typically exhibit a dependency on a single source. Specifically, the majority of models still rely predominantly on self-reflection for improvement, often treating peer information as noise or distraction. Only select high-capability agents (Claude and Qwen) demonstrate the ability to effectively leverage peer-learning to achieve performance breakthroughs. This lack of synthesis capability points to a key direction for future research: developing mechanisms that enable agents to robustly reconcile internal diagnostics with external knowledge.

\begin{figure}
    \centering
    \includegraphics[width=\linewidth]{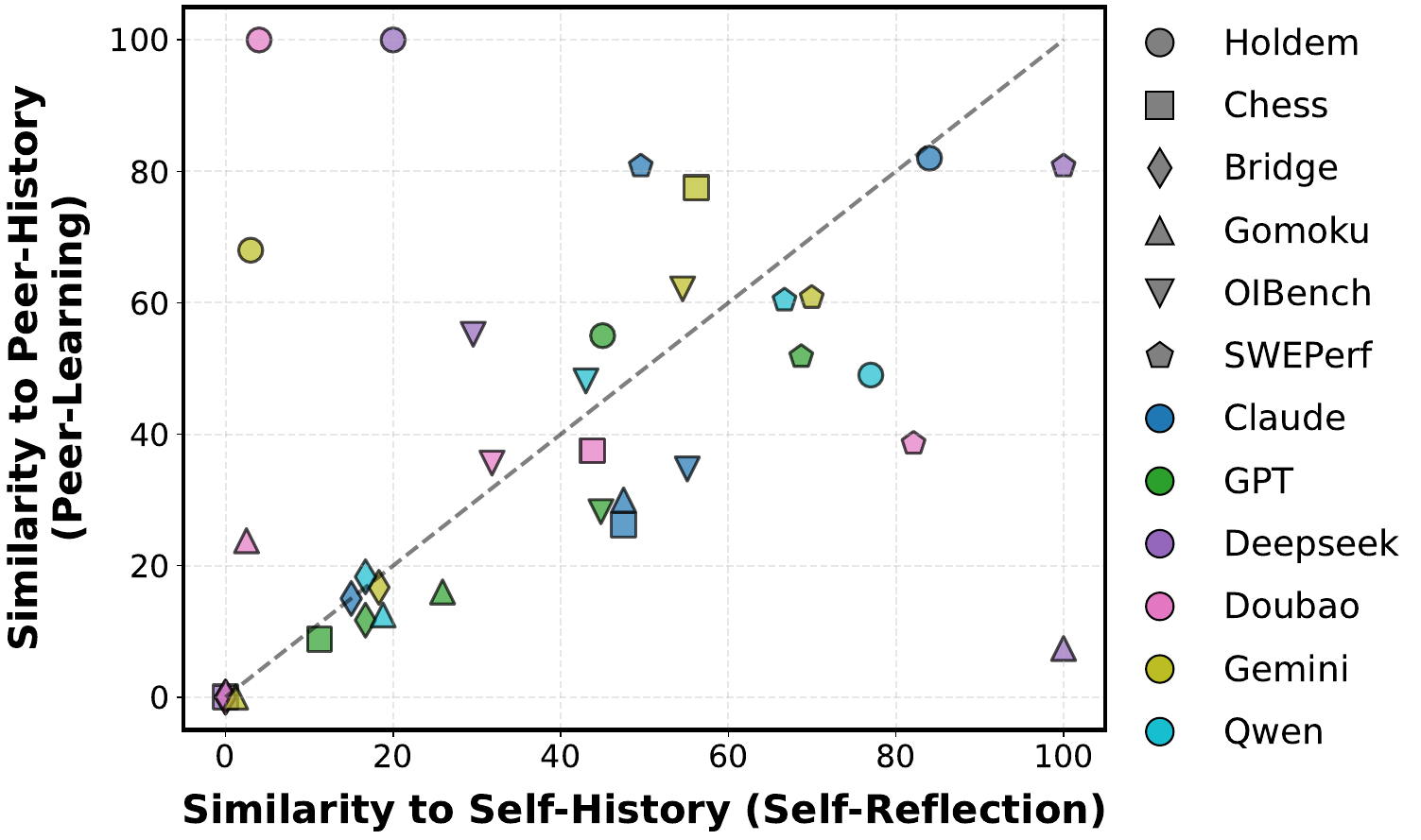}
    \caption{\textbf{Code Evolution Patterns.} Each point represents an agent's Round 2 submission. The X-axis measures similarity to its own Round 1 code, while the Y-axis measures the maximum similarity to any peer's Round 1 code.}
    \label{fig:evolution_mechanism}
\end{figure}

\textbf{Code Consistency Analysis.} To further dissect the behavioral mechanisms underlying evolution, we quantify the extent to which agents rely on self-refinement versus peer-adoption. For each agent's Round 2 submission, we compute its similarity to its own Round 1 code (X-axis, self-consistency) and the maximum similarity to any peer's Round 1 code (Y-axis, peer-adoption). For competitive tasks, similarity is measured by behavioral alignment in endgame positions; for objective tasks, we employ CodeBLEU~\cite{ren2020codebleumethodautomaticevaluation} to assess syntactic and semantic overlap. As illustrated in Figure~\ref{fig:evolution_mechanism}, most points cluster near the diagonal, indicating a balanced use of both mechanisms. However, three extreme failure modes emerge: (1) \textit{Blind Copying} (high $y$), where Doubao and DeepSeek in Hold'em completely abandon their logic to mimic peers; (2) \textit{Stubborn Stagnation} (high $x$), exemplified by DeepSeek in Gomoku, which refuses to adapt despite poor performance; and (3) \textit{Chaotic Reconstruction} ($x,y < 20$), where agents discard both internal and external priors, leading to unstable solutions. These extremes highlight a critical inability of current models to dynamically balance exploration and exploitation. To provide a more granular view of these behavioral patterns, we present the complete pairwise similarity matrices for all tasks in Appendix~\ref{app:similarity_matrices}.

\textbf{Case Study.} Development logs reveal a structured refinement process driven by peer-learning and self-reflection (See Appendix~\ref{app:casestudy}). Agents actively utilize tools such as \texttt{Grep} to extract win statistics from tournament reports, identifying top-performing peers to study and synthesize their logic. For instance, in Texas Hold'em, Claude transitioned from a basic heuristic formula to a sophisticated opponent-aware strategy after analyzing the aggressive betting patterns of high-ranking competitors. This peer-learning is complemented by explicit self-reflection, where agents benchmark their own win rates against rivals and plan targeted improvements in modules such as risk management or search depth. These behaviors show that agents evolve through deliberate optimization rather than stochastic mutations.

\subsection{Robustness and Extensibility of CATArena (RQ4)}
\label{sec:rq4}

\textbf{Generalization to Unseen Variants.}
To verify that CATArena evaluates genuine adaptive capabilities rather than the retrieval of memorized solutions, we conducted experiments on unseen variants for five tasks. These variants introduce rule perturbations, such as inverting hand rankings in Hold'em, that invalidate standard strategies present in the agents' pre-training data. Detailed results are provided in Appendix~\ref{app:variants}. As shown in Table~\ref{tab:variant}, while the variance in zero-shot base scores $S_\text{base}$ increases significantly due to non-robust agents failing to generate valid strategies, the evolutionary scores $S_\text{evo}$ maintain their discriminative power. This consistency confirms that CATArena robustly measures the intrinsic evolutionary ability of agents, independent of data contamination or specific task familiarity.

\textbf{Extensibility to Diverse Tracks.}
CATArena is designed as a modular framework, allowing seamless integration of new environments and evaluation tracks. To demonstrate this extensibility, we incorporated three additional experimental tracks: a Multilingual Track (Python, C++, Java, JS, Go), a Machine Learning (ML) Track, and a Complex Task Track (Pommerman). Detailed results are provided in Appendices~\ref{app:multilingual}, \ref{app:ML}, and \ref{app:pommerman}, respectively. We observe that advanced tracks pose significant challenges for SoTA agents. Most agents exhibit low $S_\text{base}$  and unstable evolutionary trajectories, similar to the oscillation patterns observed in Chess. This suggests that while the framework is ready for broader evaluation, current models struggle to generate robust starting points for evolution in these domains. As agent capabilities advance, CATArena will scale to these harder tasks to continue evaluating the frontier agents.

%% file: tables/RQ1.tex
\definecolor{lightgreen}{RGB}{200,255,200}  
\definecolor{lightred}{RGB}{255,200,200}    
\definecolor{lightgray}{RGB}{230,230,230}   
\newcommand{\posval}[1]{\cellcolor{lightgreen}#1}
\newcommand{\negval}[1]{\cellcolor{lightred}#1}
\newcommand{\zeroval}[1]{\cellcolor{lightgray}#1}

\begin{table*}[t]
\caption{\textbf{Main Results on CATArena.} We report the Base Performance ($S_\text{base}$) and Evolutionary Capability ($S_\text{evo}$) of Minimal and Commercial Agents across six tasks. Best results in each column are \textbf{bolded}. \textit{Note: Commercial agents are excluded from SWE-Perf due to incompatibility with the Docker environment.}}
\label{tab:rq1}
\centering
\footnotesize  
\resizebox{\textwidth}{!}{%
\setlength{\tabcolsep}{6pt}  
\renewcommand{\arraystretch}{1.0}  
\begin{tabular}{l|l|cc|cc|cc|cc|cc|cc}
\toprule
\multicolumn{2}{c|}{\multirow{3}{*}{\textbf{Agents}}} & \multicolumn{8}{c|}{\textbf{Competitive}} & \multicolumn{4}{c}{\textbf{Objective}} \\
\cmidrule(lr){3-10} \cmidrule(lr){11-14}
\multicolumn{2}{c|}{} & \multicolumn{2}{c|}{\textbf{Gomoku}} & \multicolumn{2}{c|}{\textbf{Hold'em}} & \multicolumn{2}{c|}{\textbf{Bridge}} & \multicolumn{2}{c|}{\textbf{Chess}} & \multicolumn{2}{c|}{\textbf{OIBench}} & \multicolumn{2}{c}{\textbf{SWE-Perf}} \\
\cmidrule(lr){3-4} \cmidrule(lr){5-6} \cmidrule(lr){7-8} \cmidrule(lr){9-10} \cmidrule(lr){11-12} \cmidrule(lr){13-14}
\multicolumn{2}{c|}{} & $S_\text{base}$ & $S_\text{evo}$ & $S_\text{base}$ & $S_\text{evo}$ & $S_\text{base}$ & $S_\text{evo}$ & $S_\text{base}$ & $S_\text{evo}$ & $S_\text{base}$ & $S_\text{evo}$ & $S_\text{base}$ & $S_\text{evo}$ \\
\midrule
\multicolumn{1}{l|}{\multirow{6}{*}{\rotatebox{90}{Minimal}}}
& Claude & \textbf{0.88} & \posval{+0.030} & \textbf{0.58} & \posval{+0.072} & 0.79 & \textbf{\posval{+0.062}} & \textbf{0.90} & \negval{-0.061} & 0.32 &  \posval{+0.040} & 0.02 & \negval{-0.0016} \\
& DeepSeek     & 0.23 & \posval{+0.014} & 0.01 & \posval{+0.006} & 0.00 & \zeroval{0.000} & 0.00 & \zeroval{0.000} & 0.10 & \negval{-0.029} & 0.02 & \negval{-0.0011} \\
& Doubao       & 0.33 & \negval{-0.046} & 0.04 & \negval{-0.011} & 0.20 & \posval{+0.013} & 0.58 & \negval{-0.214} & \textbf{0.41} & \posval{+0.041} & 0.00 & \posval{+0.0002} \\
& Gemini    & 0.25 &  \textbf{\posval{+0.156}} & 0.01 & \posval{+0.024} & \textbf{0.90} & \negval{-0.083} & 0.58 & \negval{-0.158} & 0.40 & \posval{+0.022} & 0.01 & \posval{+0.0012} \\
& GPT          & 0.48 & \posval{+0.012} & 0.16 & \posval{+0.025} & 0.47 & \negval{-0.095} & 0.38 & \negval{-0.044} & \textbf{0.41} & \posval{+0.031} & 0.02 & \posval{+0.0005} \\
& Qwen3    & 0.85 & \negval{-0.196} & 0.20 & \posval{+0.019} & 0.65 & \posval{+0.003} & 0.58 & \negval{-0.187} & 0.10 & \textbf{\posval{+0.099}} & 0.01 & \textbf{\posval{+0.0036}} \\
\midrule
\multirow{5}{*}{\rotatebox{90}{Commercial}} 
& Best ADK Agent    & 0.00 & \textbf{\posval{+0.176}} & 0.07 & \textbf{\posval{+0.057}} & 0.25 & \textbf{\posval{+0.191}} & \textbf{0.91} & \negval{-0.072} & 0.45 & \posval{+0.049} & - & - \\
& Claude Code       & 0.78 & \negval{-0.035} & 0.01 & \posval{+0.027} & \textbf{1.00} & \posval{+0.056} & 0.56 & \negval{-0.178} & 0.43 & \posval{+0.072} & - & - \\
& Codex   & 0.47 & \posval{+0.103} & \textbf{0.72} & \negval{-0.008} & 0.75 & \posval{+0.062} & 0.38 & \posval{+0.011} & 0.46 & \textbf{\posval{+0.186}} & - & - \\
& Gemini CLI        & 0.31 & \negval{-0.030} & 0.13 & \negval{-0.017} & 0.01 & \negval{-0.056} & 0.38 & \textbf{\posval{+0.099}} & \textbf{0.63} & \negval{-0.026} & - & - \\
& Qwen Code  & \textbf{0.94} & \posval{+0.035} & 0.07 & \posval{+0.025} & 0.49 & \posval{+0.053} & 0.28 & \negval{-0.144} & 0.35 & \posval{+0.009} & - & - \\
\bottomrule
\end{tabular}
}
\vspace{-8pt}
\end{table*}

%% file: tables/task.tex
\begin{table}[t]
\centering
\scriptsize
\caption{\textbf{Evolutionary Population Dynamics.} Pearson correlation coefficients between iteration rounds and population performance metrics: $\text{DIS}_\text{std}$ (Standard Deviation) and $\text{Trend}_\text{mean}$ (Mean Score).}
\label{tab:dynamics}
\setlength{\tabcolsep}{4pt}
\begin{tabular}{llccc}
\toprule
\textbf{Task Type} & \textbf{Task} & $\text{DIS}_\text{std}$ & $\text{Trend}_\text{mean}$ & \textbf{Pattern} \\ 
\midrule
\multirow{2}{*}{Objective} 
  & OIBench   & +0.42 & +0.79 & Divergence \\
  & SWE-Perf  & +0.20 & +0.20 & Weak trend \\
\midrule
\multirow{4}{*}{Competitive}
  & Gomoku    & -0.05 & +0.42 & Convergence \\
  & Hold'em   & -0.81  & +0.75 & Strong convergence \\
  & Bridge    & -0.82 & +0.24 & Moderate convergence \\
  & Chess     & +0.55 & -0.74 & Oscillation \\
\bottomrule
\end{tabular}
\end{table}

%% file: tables/ablation_PL.tex
\begin{table}[t]
\centering
\caption{\textbf{Ablation Study on peer-learning.} Comparison of performance with full feedback ($G_2$) versus self-reflection only ($G_2'$). \textbf{Bold} values indicate improvement over the initial round ($G_1$). The \textit{Pattern} column denotes the effective mechanism: S indicates effective Self-Reflection ($G_2' > G_1$), P indicates effective Peer-Learning ($G_2 > G_2'$).}

\label{tab:ablation_PL}
\footnotesize  
\resizebox{\linewidth}{!}{%
\setlength{\tabcolsep}{4pt}  
\renewcommand{\arraystretch}{1.0}  
\begin{tabular}{l|cccc|cccc}
\toprule
\multirow{2}{*}{\textbf{Agent}} & \multicolumn{4}{c|}{\textbf{Gomoku}} & \multicolumn{4}{c}{\textbf{OIBench}} \\
\cmidrule(lr){2-5} \cmidrule(lr){6-9}
 & $G_1$ & $G_2'$ & $G_2$ & {Pattern} &  $G_1$ & $G_2'$ & $G_2$ & {Pattern} \\
\midrule
Claude & 0.814 & 0.520 & \textbf{0.951} & \ding{55}P & 0.318 & 0.204 & \textbf{0.410} & \ding{55}P \\
DeepSeek & 0.363 & 0.265 & 0.333 & \ding{55}\ding{55} & 0.097 & \textbf{0.105} & 0.000 & S\ding{55} \\
Doubao & 0.716 & 0.039 & 0.598 & \ding{55}\ding{55} & 0.408 & \textbf{0.458} & 0.000 & S\ding{55} \\
Gemini & 0.225 & \textbf{0.794} & 0.118 & S\ding{55} & 0.403 & 0.346 & 0.301 & \ding{55}\ding{55} \\
GPT & 0.500 & 0.480 & 0.020 & \ding{55}\ding{55} & 0.413 & \textbf{0.470} & \textbf{0.423} & S\ding{55} \\
Qwen & 0.833 & 0.814 & 0.618 & \ding{55}\ding{55} & 0.101 & 0.084 & \textbf{0.398} & \ding{55}P \\
\bottomrule
\end{tabular}
}
\vspace{-8pt}
\end{table}

%% file: section/Conclusion.tex
\section{Conclusion}
In this paper, we present CATArena, a comprehensive framework that shifts the evaluation of code agents from static correctness to dynamic evolutionary capability. By simulating the iterative nature of real-world software engineering within a competitive tournament structure, CATArena quantifies the ability of agents to leverage execution feedback and peer knowledge for continuous optimization. Through extensive evaluations on CATArena, we benchmark the evolutionary capabilities of state-of-the-art LLM code agents, confirming that evolutionary potential is a distinct dimension from single-turn generation capability. Our in-depth analysis further dissects these evolutionary patterns, revealing the specific contributions of self-reflection and peer-learning mechanisms to agent growth. Moreover, we demonstrate that evolutionary trajectories are highly task-dependent; CATArena’s diverse spectrum of competitive and objective tasks proves essential for capturing these nuanced behavioral shifts. In the future, CATArena will integrate more tasks and serve as a highly extensible testbed.

%% file: section/Appendix.tex
\section{Evaluation Metric Calculation}\label{app:W_calc}
\input{tables/evaluationmetric}

To provide a unified evaluation standard across diverse tasks, we construct a global performance matrix $W$. Scoring rules for the tasks are summarized in Table~\ref{tab:game_scoring}.
Let $N$ be the number of rounds and $M$ be the number of participating agents. 
We construct a matchup matrix
\[
W \in \mathbb{R}^{(N\cdot M)\times (N\cdot M)},
\]
where the generic element $W_{i,j}^{n,m}$ denotes the score when agent $i$ from round $n$ plays against agent $j$ from round $m$. 
We abbreviate $W^n_{i,j}$ for same-round comparisons ($n=m$) and $W_i^{n,m}$ for self-comparisons across rounds ($i=j$). 
Diagonal entries $(n,i)=(m,j)$ are ignored. 

For Hold'em, pairwise results are not feasible; instead, batch-based tournaments are used and the score matrix records the win rates of multi-agent matches. For each batch, we obtain a single score group $W_{i_1, i_2, \ldots, i_{BS}}^{n_1, n_2, \ldots, n_{BS}}$, where $BS$ is the batch size. We conduct three types of experiments to accommodate different metric calculations: 
(1) $W_{i}^{1,2,\ldots,N}$, where the same agent's strategies from $N$ rounds compete against each other, used to compute self-improvement metrics $S_i^n$; 
(2) $W_{i_1, i_2, \ldots, i_T}^{n}$, where all agents in the same round compete, used to calculate the base score $B$; 
(3) All $N \times T$ agent strategies are randomly shuffled and grouped for competition (with $BS=12$ in our experiments), used to compute both the global score $G_i^n$ and advanced score $A_i^n$.

For objective tasks, we simplify the notation $W_{i,j}^{n,m}$ to $W_i^n$.

\section{Tournament Format and Repetition}\label{app:tournamentFormat}

\input{tables/TournamentConfig}

\begin{figure}[htb]
    \centering
    \includegraphics[width=0.6\linewidth]{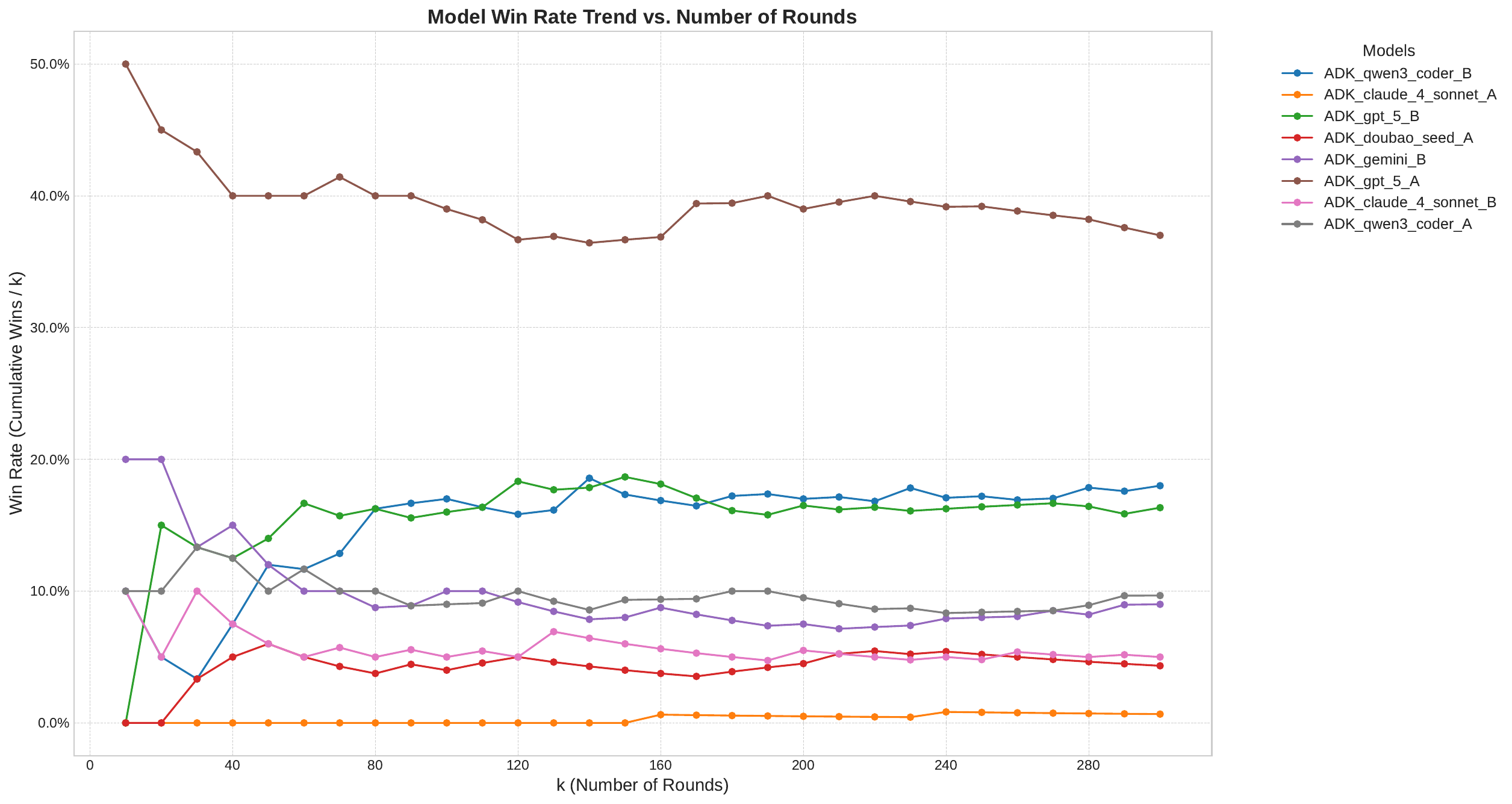}
    \caption{Cumulative win rate trajectories of Hold'em agents over 300 tournament rounds.}
    \label{fig:winrate}
\end{figure}

We list the basic settings for each task in Table~\ref{tab:game_config}.

For competitive tasks, we ensure that the number of pairwise matches among code agents allows the final results to stabilize, i.e., for each task, the L1-norm fluctuation of the scoring matrix $W$ is less than 5\%. 

For objective tasks, agents are evaluated independently on the same fixed benchmark each round (no pairwise matches). 
Each agent submits code, the arena runs the official evaluator to compute $W_i^n$, 
and a round leaderboard is produced. The leaderboard, logs, and top solutions/patches are fed back to all agents for the next round, 
forming an iterative optimization tournament over $N$ rounds with a diagonal score matrix.

Specifically, due to the inherent randomness in card dealing in Texas Hold'em, we conduct an additional analysis on the stability of the win rates.   Figure~\ref{fig:winrate} illustrates the cumulative win rate trends over 10 to 300 rounds for a representative tournament.  To further assess the reliability of our evaluation, we analyze the win rate trends from different starting rounds to the final round. Based on this analysis, starting from round 50, the cumulative win rates of all models become stable, with the absolute value of the linear regression slope dropping below 0.0002 (i.e., per-round changes less than 0.02\%). This indicates that our tournament design of 100 rounds, with each round consisting of 200 to 720 hands, is sufficient to cover the stochasticity of the game and provides a robust and reliable assessment of model performance.
The results of the tournament reliably reflect the true capabilities of the evaluated models, ensuring the soundness of subsequent algorithm comparisons and optimizations.


\section{Comparative Analysis of Elo Rating and Global Win Rate}\label{app:elo_vs_g}

\begin{figure}[htb]
    \centering
   \begin{subfigure}[b]{0.48\textwidth}
   \centering
\includegraphics[width=\linewidth]{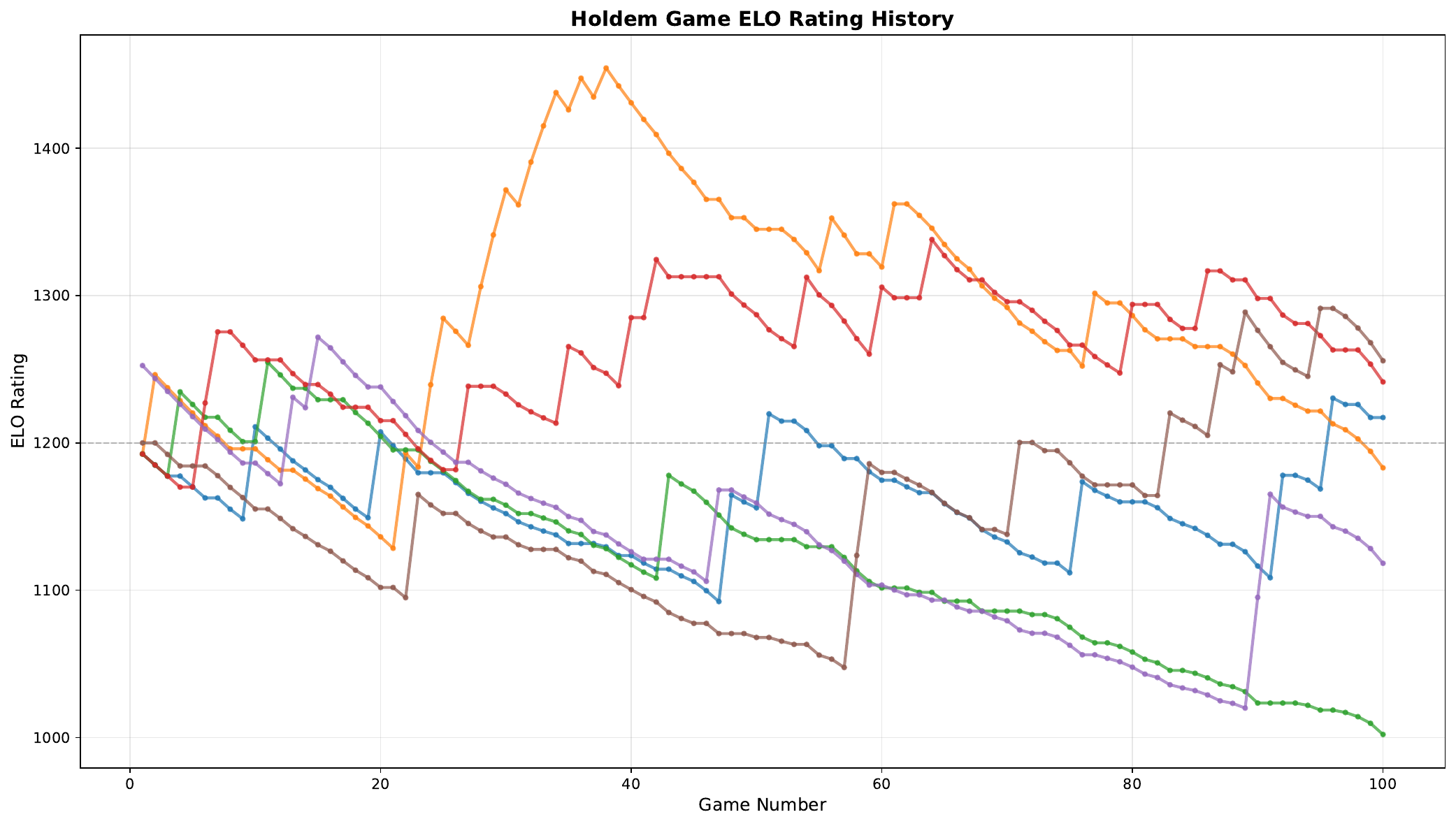}
\caption{Elo Rating History}
   \end{subfigure}
\hfill
   \begin{subfigure}[b]{0.48\textwidth}
   \centering
\includegraphics[width=\linewidth]{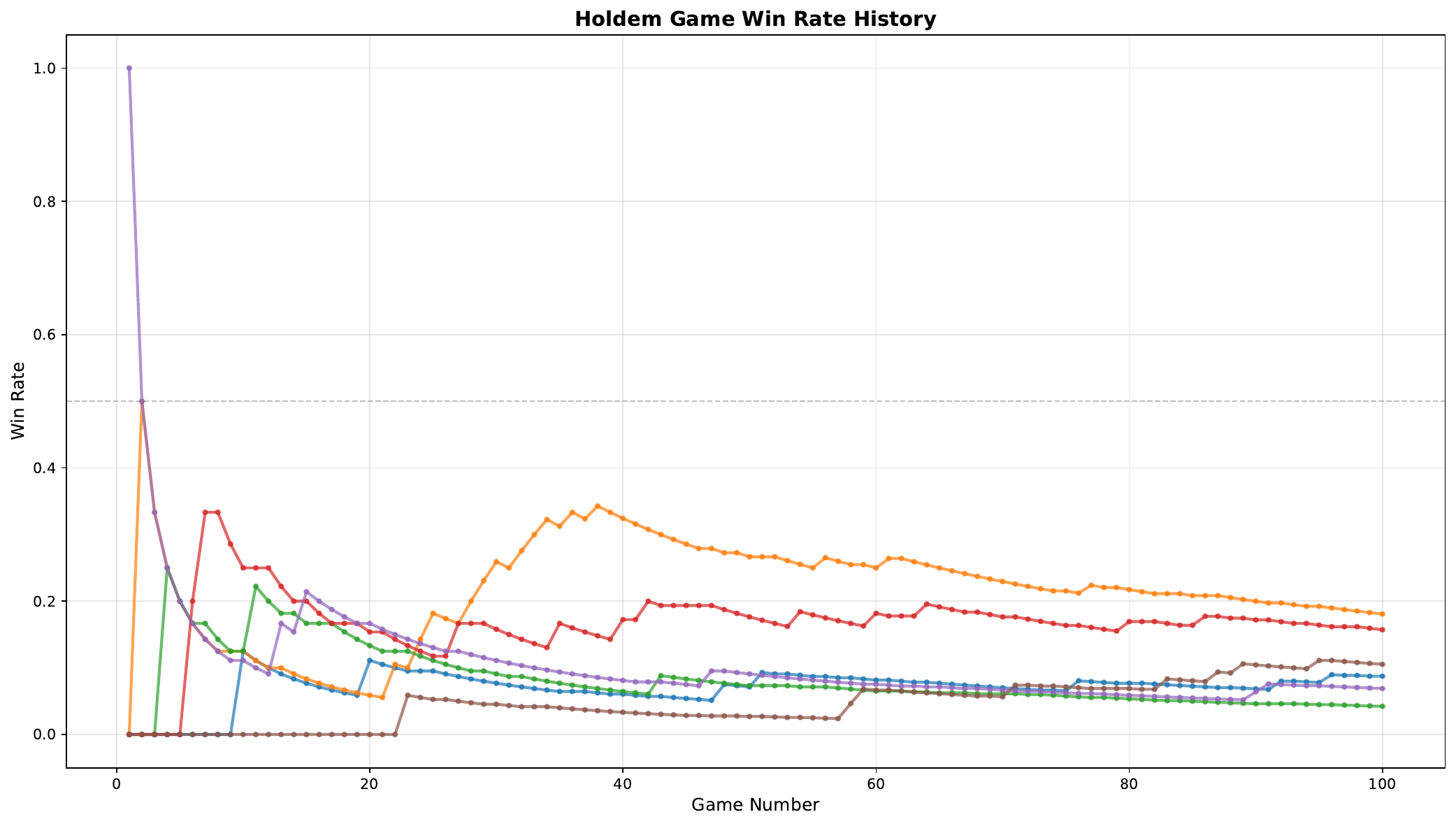}
\caption{Win Rate History}
   \end{subfigure}
   \caption{\textbf{Comparison of performance metrics in a 100-game Texas Hold'em tournament among 6 random selected players.} (a) The Elo rating fluctuates significantly due to match order sensitivity. (b) The global win rate converges to a stable value, providing a more reliable foundation for calculating the evolutionary growth rate.}
   \label{fig:elo_vs_winrate}
\end{figure}

In this section, we justify our choice of the global average win rate (the basis for our $G$ score) over the Elo rating system by comparing their stability in a 100-game Texas Hold'em tournament. As visualized in Figure~\ref{fig:elo_vs_winrate}, the Elo rating (a) exhibits significant volatility and sensitivity to match order, characterized by "zigzag" oscillations that reflect transient outcomes rather than intrinsic capability. This "recency bias" makes Elo unsuitable for capturing a stable performance snapshot for each evolutionary round. In contrast, the Global Win Rate (b) demonstrates smooth convergence and effectively filters out stochastic noise. By reflecting the agent's expected performance against the entire population, the win rate provides a robust and consistent metric, ensuring that the calculated evolutionary slope represents genuine strategic growth rather than artifacts of tournament scheduling.

\section{Generation Configs}\label{app:generationconfig}
\input{tables/AgentDescription}

For all LLMs used in our work, we set the temperature to 0.1 and the max tokens to the official API defaults. We set top-p to 1.0, top-k to 100, and the presence penalty to the API default.

Additionally, both Claude-4-Sonnet and DeepSeek-3.1 occasionally encounter tool call issues that result in no code being generated, as frequently reported by the community. If such errors occur three times in a row, we substitute Claude-4-Sonnet with Claude-3.7-Sonnet and DeepSeek-3.1 with DeepSeek v3.


\section{Repetition Experiments: Stability Analysis}
\label{sec:repeat_exp}

\input{tables/repeatR1}

Table~\ref{tab:std_rounds} presents the results of four independent repetitions of the first two tournament rounds. The primary objective is to assess whether stochasticity in code generation significantly alters the competitive landscape.

The data reveals that ranking stability is robust: \textbf{the majority of agents maintain their relative positions with a standard deviation of less than one rank.} While specific implementations vary per run, the performance gap between agents is sufficiently large to maintain a stable leaderboard.

Other minor observations include: (1) standard games exhibit higher stability than variant games; (2) open-source and commercial agents generally outperform minimal agents in terms of consistency. Finally, we observe that "runnability" is not guaranteed across runs, suggesting that while the relative capability (ranking) is stable, the absolute success rate of code generation still suffers from inherent randomness.

\section{Error Rate Analysis}\label{app:httpError}

\input{tables/errorRate}

 During the implementation of strategy code, some agents encounter runtime errors. We observe that all agents were able to generate the required files successfully. However, most errors occur in interface implementation and action validity. These cases were consistently recorded as HTTP errors.  Whenever an HTTP error occurs, a default action is automatically executed to ensure the continuity of the competition. For Texas Hold'em Poker, the default action is to fold for that hand. For Chess and Gomoku, the default action is for the player to concede the current game.

Table~\ref{tab:errorRate} reports the proportion of HTTP error actions encountered during the code development of each agent in Round 1.  The error rate is calculated as the rate of failed actions in total number of actions taken by the strategy code.  The experimental results indicate that commercial agents exhibit a significantly lower failure rate compared to minimal agents. Additionally, the failure rate is higher in variant games than in traditional games.

\section{Detailed Evolutionary Trajectory Analysis}\label{app:learning}

\begin{figure}[htb]
    \centering
    \includegraphics[width=\linewidth]{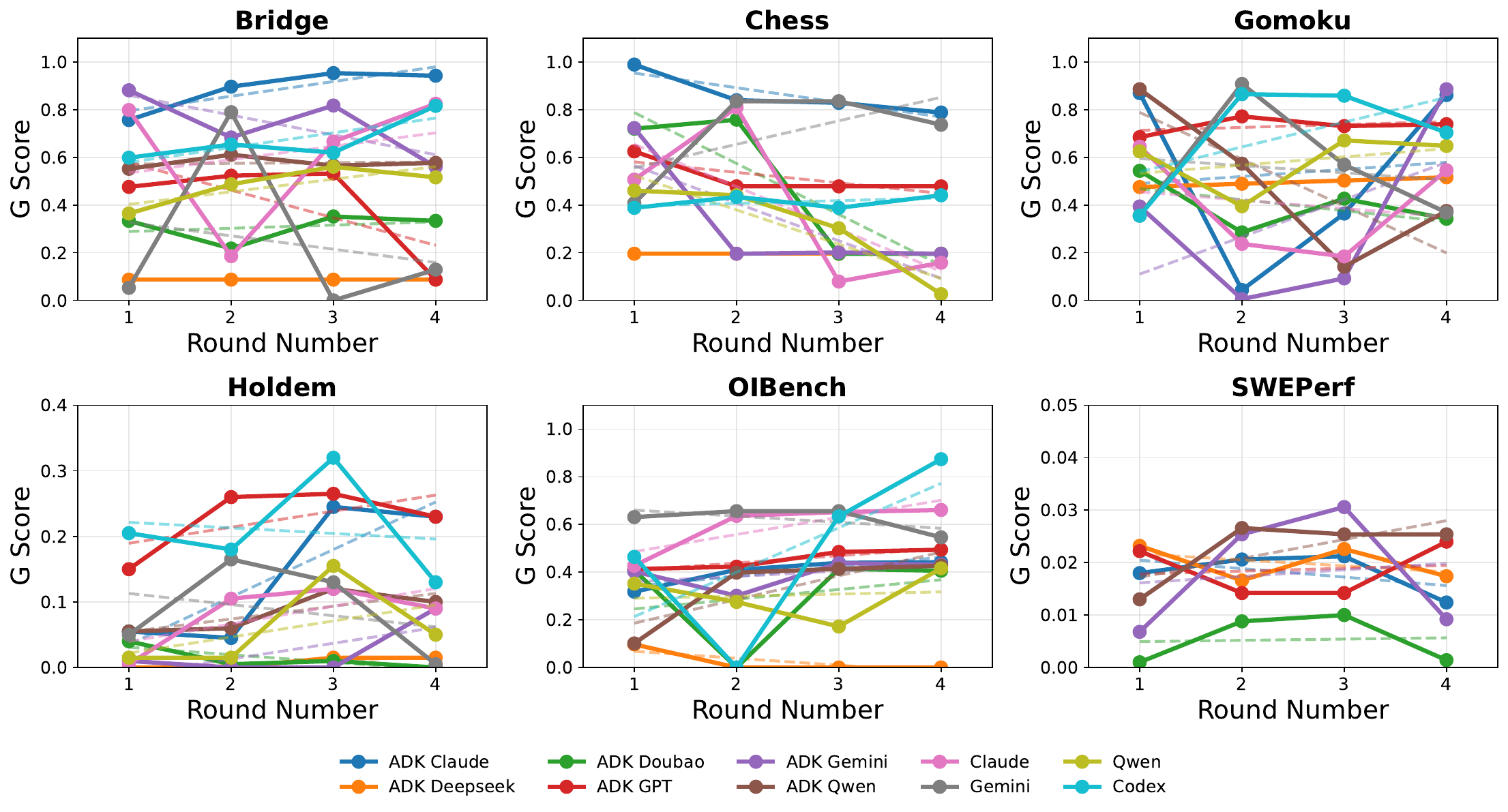}
    \caption{Trend of global performance score $G_i^n$ in all tasks.}
    \label{fig:learning_figure}
\end{figure}

In this section, we provide a comprehensive visualization of the evolutionary trajectories across all six tasks. Figure~\ref{fig:evolution_trajectory} plots the round-wise global performance $G^n$ for each agent, with dashed lines representing the fitted regression of $S_\text{evo}$ to reflect the evolution trend.

\begin{enumerate}[leftmargin=10pt]
    \item \textbf{Universality of Performance Oscillation.} The visualization confirms that performance volatility is not task-specific but a universal phenomenon in agent-based evolution. In environments with high strategic complexity, such as Gomoku and Bridge, many agents (e.g., ADK Claude, ADK Gemini) experience significant performance drops in intermediate rounds before potentially recovering. This widespread oscillation reinforces our conclusion in the main text: current agents often fail to reliably distinguish between beneficial and detrimental strategy modifications, leading to the accidental discarding of high-performing code versions.
    \item \textbf{Latent Potential Discovery via Iteration.} The trajectories across different tasks suggest that static, zero-shot evaluations often fail to capture an agent's upper bound. Even in the highly challenging SWE-Perf task, despite the low absolute scores, certain agents exhibit clear upward trajectories, suggesting that iterative feedback allows agents to gradually navigate the complex search space of software engineering.
    \item \textbf{Reliability of the $S_\text{evo}$ Metric.} The figure highlights the importance of our proposed $S_\text{evo}$. While the raw $G^n$ is highly sensitive and sometimes suffers from development failures in specific rounds, the dashed regression lines remain relatively stable. This indicates that our evolutionary score provides a more accurate measure of the agent's evolutionary progress than any single-round performance snapshot.
\end{enumerate}

\section{Pairwise Code Similarity Heatmaps}\label{app:similarity_matrices}
\begin{figure}[htb]
    \centering
    \begin{subfigure}[b]{0.32\textwidth}
        \centering
        \includegraphics[width=\linewidth]{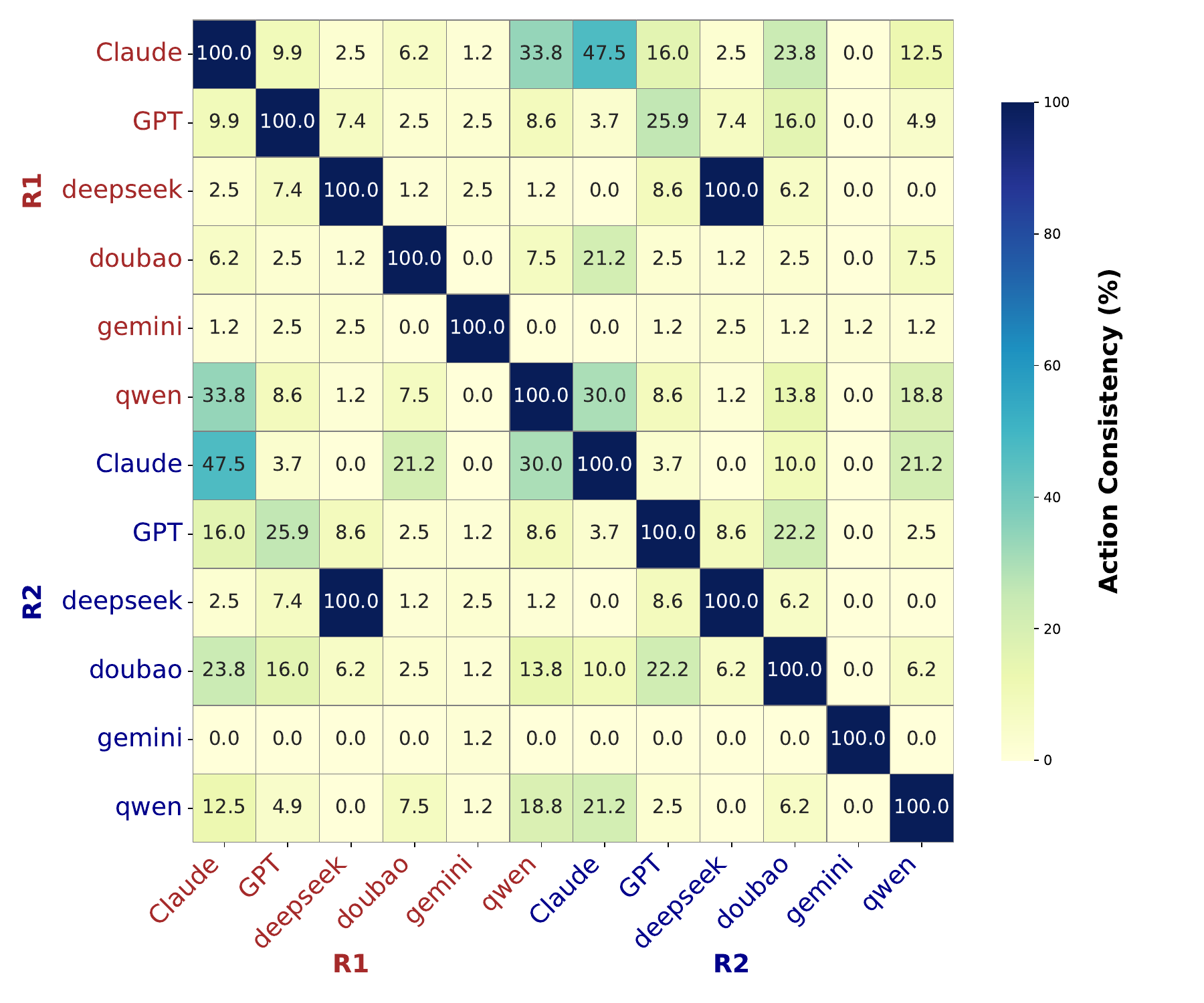}
        \caption{Gomoku.}
    \end{subfigure}
    \hfill
    \begin{subfigure}[b]{0.32\textwidth}
         \centering
        \includegraphics[width=\linewidth]{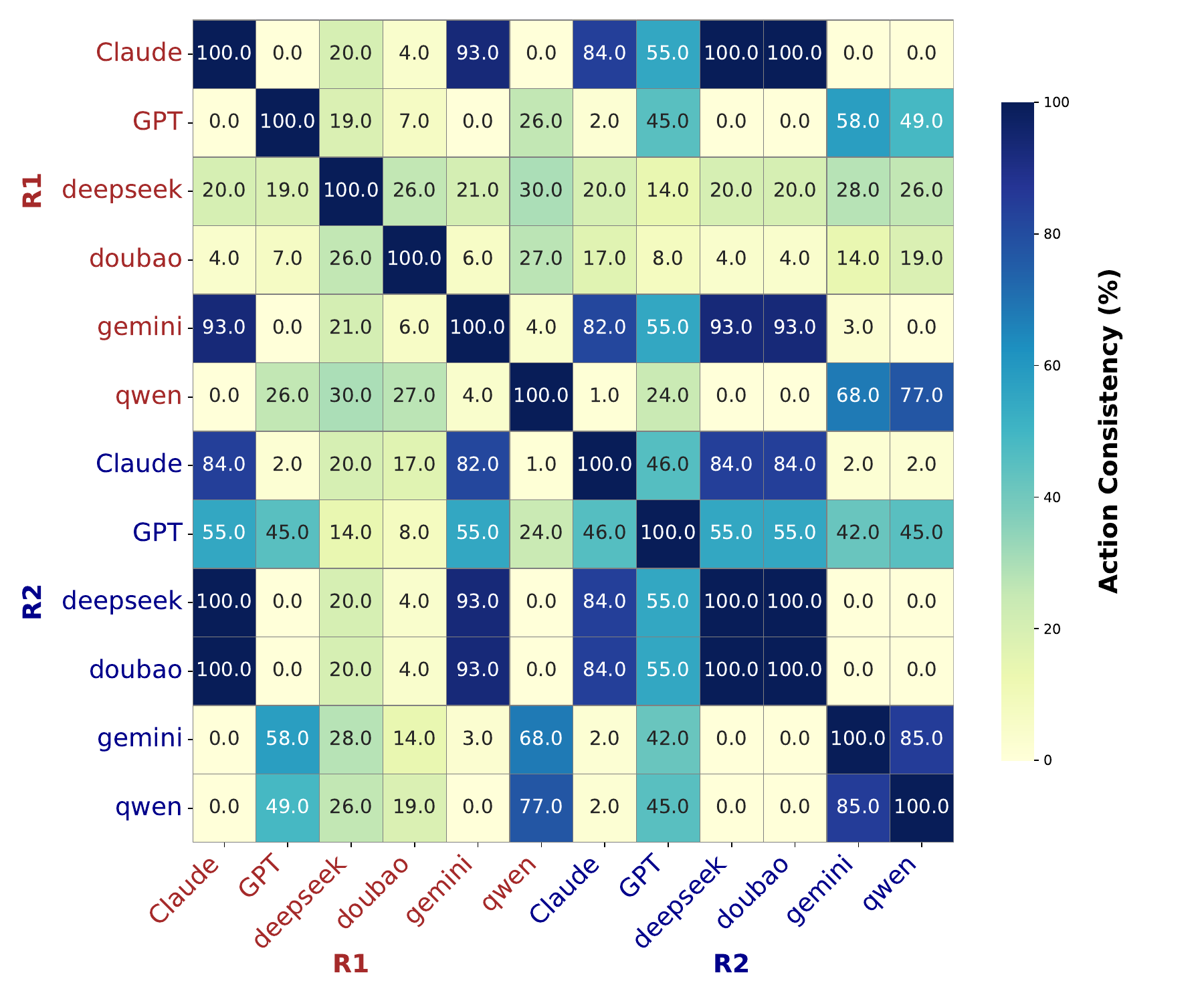}
        \caption{Hold'em.}
    \end{subfigure}
    \hfill
     \centering
       \begin{subfigure}[b]{0.32\textwidth}
       \centering
            \includegraphics[width=\linewidth]{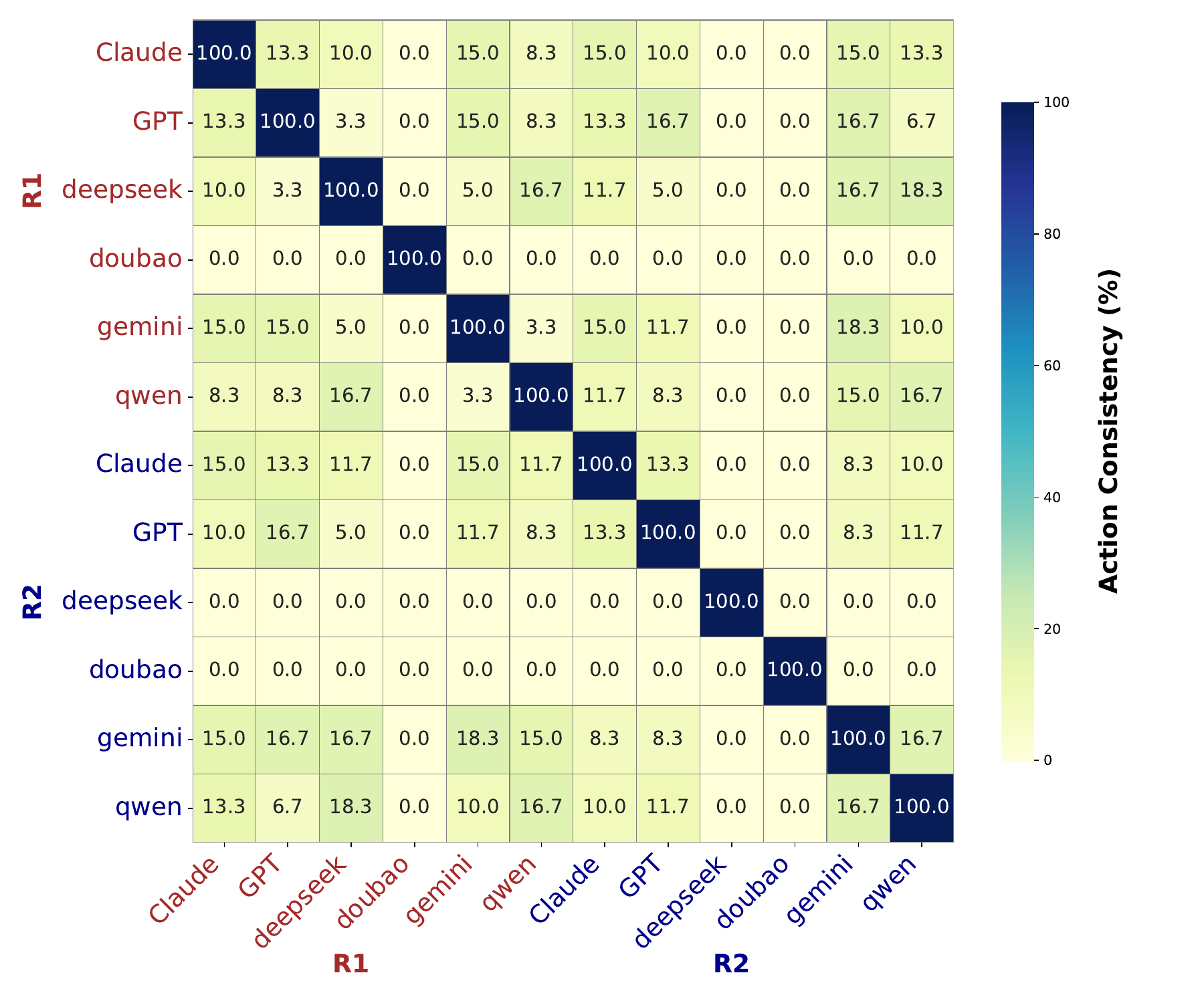}
        \caption{Bridge.}
       \end{subfigure}

       \vspace{0.5em}

       \begin{subfigure}[b]{0.32\textwidth}
            \includegraphics[width=\linewidth]{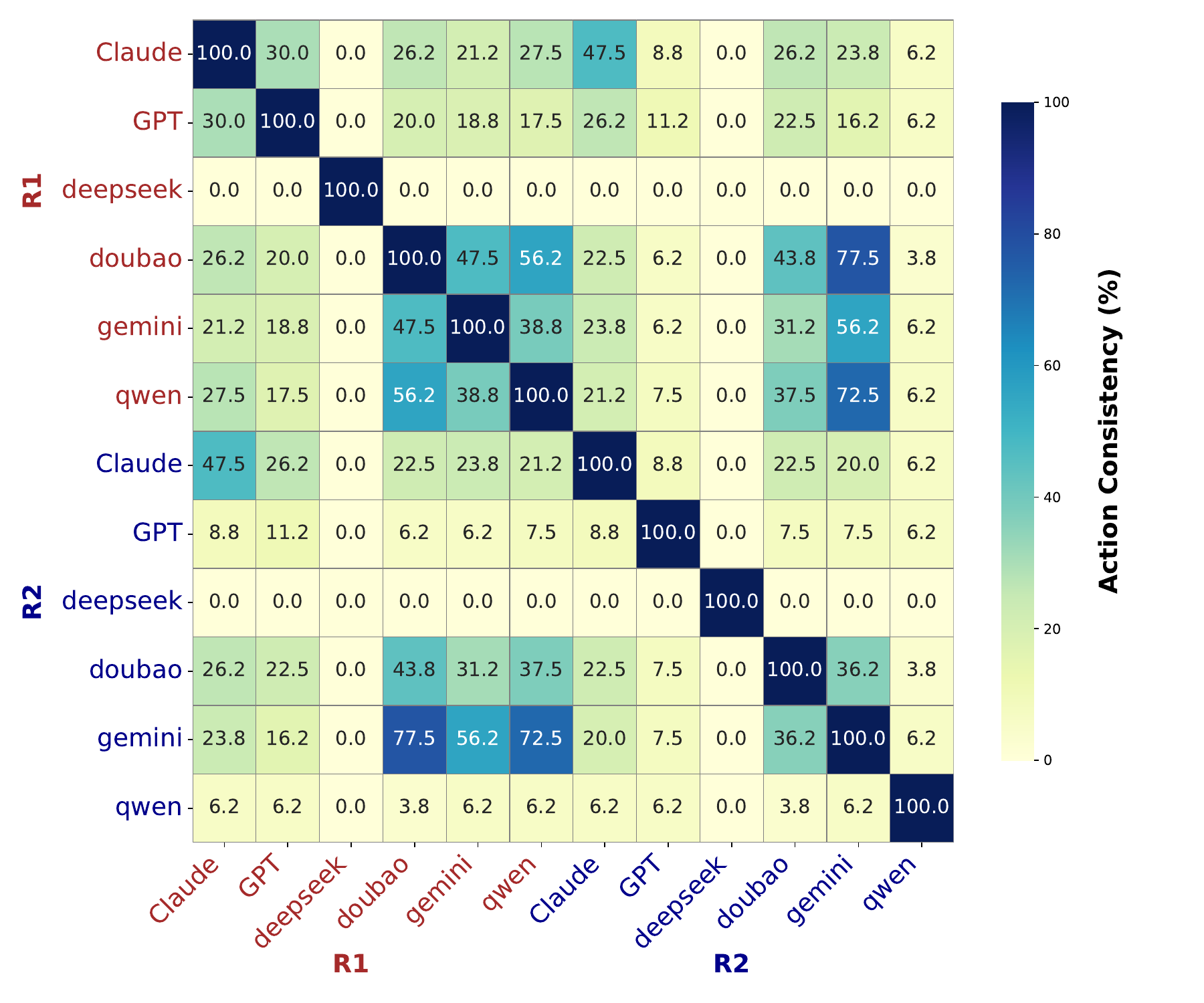}
        \caption{Chess.}
       \end{subfigure}
       \hfill
       \begin{subfigure}[b]{0.32\textwidth}
            \includegraphics[width=\linewidth]{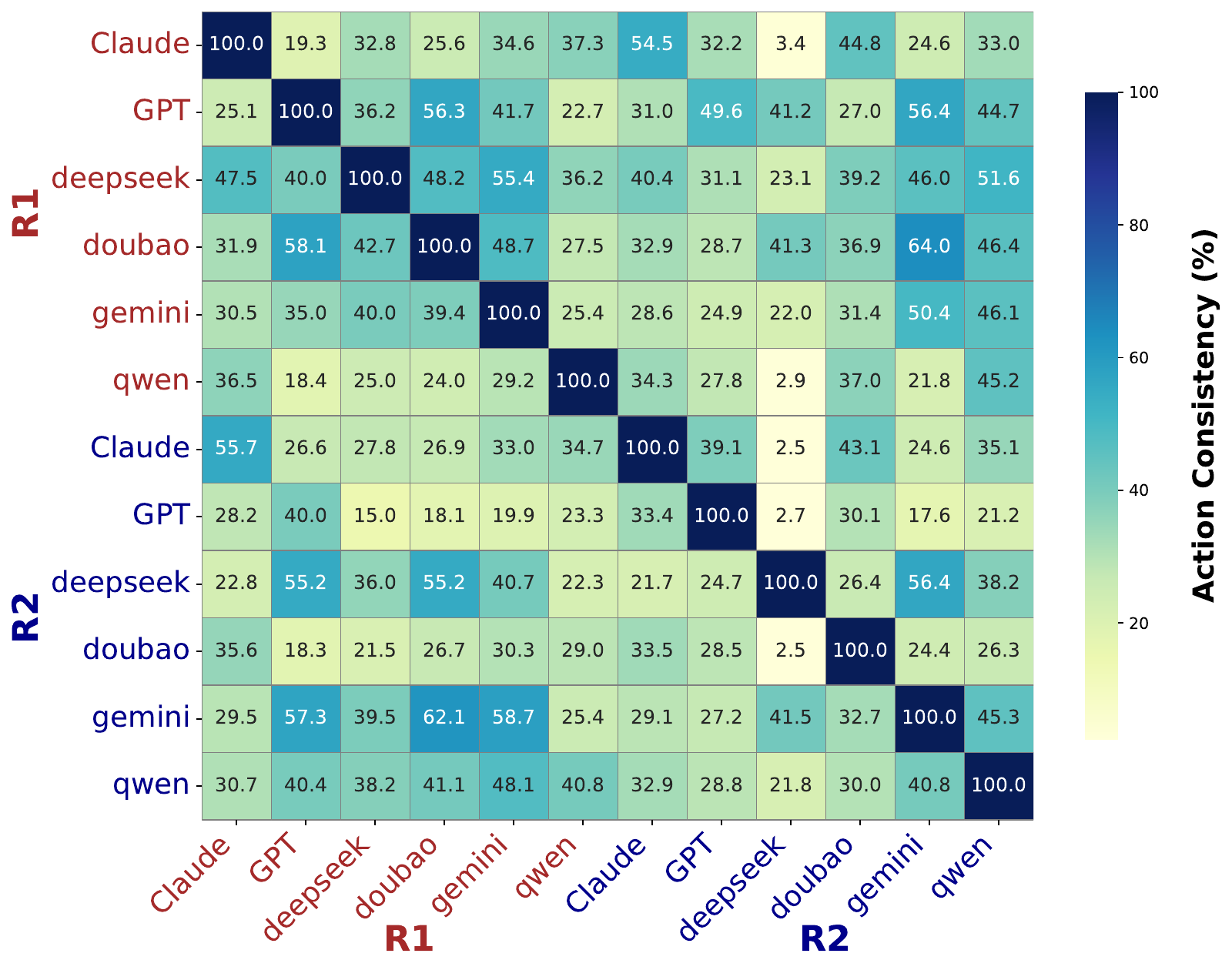}
        \caption{OIBench.}
       \end{subfigure}
       \hfill
       \begin{subfigure}[b]{0.32\textwidth}
            \includegraphics[width=\linewidth]{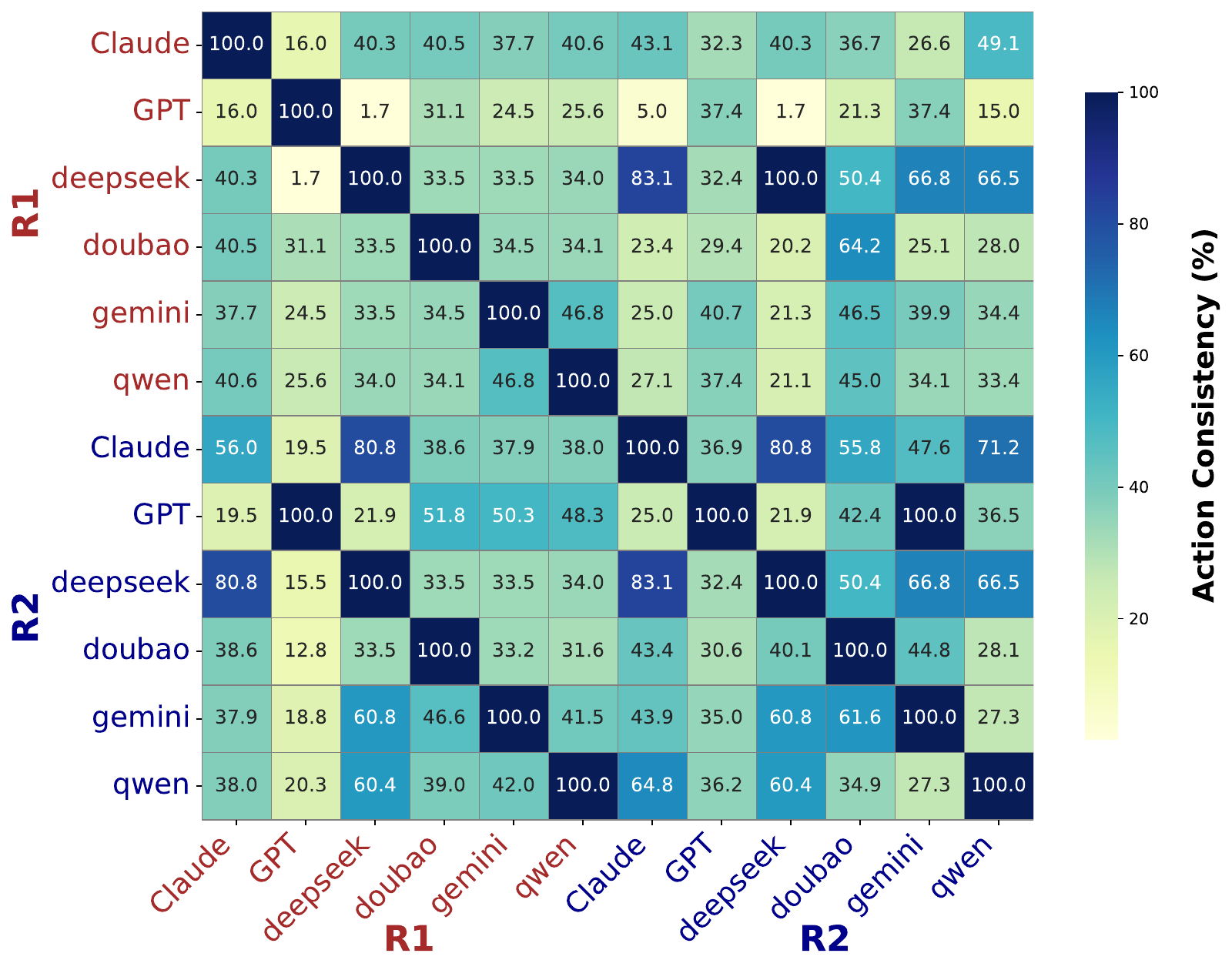}
        \caption{SWE-Perf.}
       \end{subfigure}
    \caption{\textbf{Pairwise similarity matrices between Round 1 and 2 code submissions across six tasks.} For competitive games (a-d), similarity is measured by Action Consistency in endgame positions. For programming tasks (e-f), similarity is measured by CodeBLEU. }
    \label{fig:similarity_heatmaps}
\end{figure}

In this section, we provide the detailed pairwise similarity matrices between Round 1 and Round 2 code submissions for all six tasks (Figure~\ref{fig:similarity_heatmaps}). In Competitive Tasks, similarity is measured by Action Consistency (\%). A value of 100.0 indicates that the agent's code produces identical behavioral outputs in endgame positions. In objective tasks, similarity is measured by CodeBLEU. These matrices generally show lower absolute values and higher dispersion, reflecting the complexity of syntactic and semantic alignment in large-scale code repositories. 

A key observation is that the distribution and range of similarity scores vary significantly by task, reflecting the inherent difficulty of learning and replicating code in different domains.
In Hold'em (b), the matrix is characterized by exceptionally high similarity values, including several instances of 100.0\% alignment. This indicates that the logic required for this task is relatively "easy to learn" and replicate. Agents can effortlessly adopt a peer's strategy or maintain their own with perfect consistency, suggesting a lower barrier to entry for strategic imitation in this environment. In Gomoku (a), Bridge (c), and Chess (d), the similarity scores are more dispersed. In objective tasks, even the highest similarity scores rarely exceed 60-70\%. This is due to two factors: (1) These tasks involve larger codebases and more intricate dependencies, making it difficult for agents to maintain high syntactic or semantic overlap during refinement. (2) The use of CodeBLEU means that even minor functional improvements or refactoring can lead to a significant drop in measured similarity.

\section{Case Study of Agent Learning} \label{app:casestudy}

To demonstrate how agents actively improve their strategies by analyzing other agents' code and tournament reports, we present a detailed case study focusing on \texttt{Claude-Code\_A} in Texas Hold'em Poker tournaments. This case clearly shows that score changes across rounds reflect the agent's understanding and adaptation to opponent strategies.

\paragraph{Development Planning and Evidence of Learning}

Analysis of the Round 3 development log shows that the agent explicitly planned to study other agents' strategies. The planning log has the following entries:
\begin{verbatim}
1. "Read game project documentation and requirements" - completed
2. "Analyze tournament rules and mode" - ongoing
3. "Study last round tournament reports and AI strategies" - pending
4. "Examine existing AI code from round 2" - pending
\end{verbatim}

\paragraph{{Tool Use Summary}}

Table~\ref{tab:tool_use_claude} summarizes all the learning-related tool use operations performed by \texttt{claude-code\_A} during the development of its Round 3 code. The tool use logs show that the agent first located and read the tournament report file, extracted win statistics using \texttt{Grep} (codex\_A: 35 wins, gemini-cli\_B: 30 wins, claude-code\_A: 10 wins), then listed all agents, and finally read the code of the top performers.

\begin{table}[htbp]
    \centering
    \small
    \caption{Summary of learning-related tool use operations by \texttt{claude-code\_A} (Round 3 development).}
      \label{tab:tool_use_claude}
    \begin{tabularx}{\textwidth}{l l X}
        \hline
      Tool & Operation & Purpose \\
        \hline
      Glob & Search \texttt{**/tournament\_report\_tourney\_*.json} & Locate tournament report files \\
        Glob & Search \texttt{**/history\_tourney\_*.json} & Locate tournament history files \\
        Read & Read \texttt{tournament\_report\_tourney\_075455.json} & Analyze agent performance \\
       Read & Read first 100 lines of report & Quickly understand report structure \\
        Grep & Search for \texttt{"wins":} field & Extract win counts for each agent \\
      Glob & List \texttt{round\_2/**/*} & Discover all agents in previous round \\
         Read & Read own previous solutions & Review existing implementation \\
        Read & Read \texttt{codex\_A} code &  Peer-strategy extraction (Top performer, 35 wins) \\
      Read & Read \texttt{gemini-cli\_B} code & Cross-agent logic synthesis (second performer, 30 wins) \\
        \hline
    \end{tabularx}
\end{table}

\paragraph{Strategy Adaptation in Code}
Based on the discovery that top-performing agents (codex\_A) utilized aggressive betting patterns, the agent transitioned from a static probability-based model to a dynamic opponent-aware strategy. Comparing the code between rounds, it is evident that the agent made targeted improvements based on its analysis:
\begin{itemize}[leftmargin=10pt]
    \item \textbf{Round 2:} Basic Chen formula evaluation, simple tournament strategy, no opponent modeling.
    \item \textbf{Round 3:} Added opponent analysis (\texttt{analyze\_opponents()}), improved decision formula incorporating opponent aggression, refined stack and position factors.
\end{itemize}

Example of newly added code in Round 3:
\begin{lstlisting}[language=Python, basicstyle=\ttfamily\small, breaklines=true]
def analyze_opponents(self, game_state: Dict[str, Any]) -> Dict[str, float]:
    """Analyze opponent tendencies based on action history"""
    action_history = game_state.get('action_history', [])
    # ... analyze aggression for each opponent ...
    return opponent_stats

# Use opponent analysis in decision making
opponent_stats = self.analyze_opponents(game_state)
avg_opponent_aggression = sum(opponent_stats.values()) / max(len(opponent_stats), 1)
decision_strength = (
    hand_strength * 
    (1 + position_factor * 0.4) * 
    stack_factor * 
    bubble_factor *
    (1 + (1 - avg_opponent_aggression) * 0.2)
)
\end{lstlisting}

These code modifications are targeted responses to the strategies employed by top-performing agents, enabling more effective adaptation and strategic improvement.

\paragraph{Reflections in Development Summary}

After completing development, the agent's summary explicitly compares its performance and highlights improvements:

\begin{tcolorbox}[colback=gray!10!white, colframe=gray!80!black, title=Agent Self-Reflection Log, sharp corners]
\textbf{Performance Analysis from Round 2}

Based on the tournament results:

- \textbf{codex\_A}: 35 wins (best performer)

- \textbf{gemini-cli\_B}: 30 wins

- \textbf{claude-code\_A}: 10 wins (my previous version)

My new AI improves on the previous version with:

- Better preflop ranges and position play

- Enhanced tournament strategy

- More sophisticated opponent modeling

- Improved risk management
\end{tcolorbox}

Similar learning behavior is observed for \texttt{codex\_A} and other agents. This case study clearly demonstrates that agents actively analyze tournament reports and top-performing code via explicit tool use operations. The sequence and content of these operations, combined with targeted code improvements and reflective summaries, confirm that score changes across rounds are a reliable indicator of the agent's growing understanding and adaptation to competitor strategies.

This micro-level behavioral loop consists of performance benchmarking, code synthesis, and self-reflection. It provides a mechanistic explanation for the macro-level performance leaps observed in our evolutionary trajectories, as agents successfully internalize superior logic from their competitive environment.

\section{Robustness Analysis on Variant Tasks}\label{app:variants}

This section details the variant settings and analyzes the impact of rule changes on agent performance. The primary objective is to demonstrate the robustness of the CATArena framework: even when agents cannot rely on memorized opening books or standard algorithms, our metric for evolutionary capability $S_\text{evo}$ remains a reliable indicator of their learning capability.

\paragraph{Variant Configurations} We introduce specific rule perturbations designed to disrupt established priors while preserving the core reasoning logic:
\begin{itemize}[leftmargin=10pt]
    \item \textbf{Gomoku:} Introduce forbidden points (e.g., double-three restrictions) to constrain the search space.
\item \textbf{Texas Hold'em:} Alter probability landscapes by removing specific cards and inverting hand rank hierarchies.
\item \textbf{Chess:} Adopt Chess960 (Fischer Random Chess) and modified movement rules to neutralize opening databases.
\item \textbf{Bridge:} Implement a "Card Exchange" mechanism between teammates to disrupt standard signaling conventions.
\item \textbf{OIBench:} Shift the optimization objective from time complexity to space complexity, requiring memory-efficient refactoring rather than speed optimization.
\end{itemize}

\input{tables/variant}
\paragraph{Results and Discussion}
Table~\ref{tab:variant} compares the performance metrics between standard and variant settings. We observe two critical phenomena:
\begin{enumerate}[leftmargin=10pt]
    \item \textbf{Widening Gap in Baseline Performance $S_\text{base}$}: Unlike standard tasks where scores are often tightly clustered, variant tasks exhibit higher variance in zero-shot rankings. This is because rule perturbations cause less capable agents to fail in generating valid or robust strategies, thereby amplifying the performance distinction between agents that rely on memorization versus those with genuine reasoning capabilities.
    \item \textbf{Consistency in Evolutionary Capability}:  Crucially, the evolutionary metric retains its effectiveness in distinguishing agent capabilities.  Despite the altered baselines, the relative evolutionary trends align with those in standard tasks (where most capable agents achieve $S_\text{evo} > 0$). This demonstrates that CATArena effectively captures the gradient of learning ability: it can reliably differentiate between agents that can adapt to new rules and those that cannot, proving that the evaluation of evolutionary potential is robust to task variations.
\end{enumerate}

\section{Results of Multilingual Track}\label{app:multilingual}
\input{tables/MulLang}

Table~\ref{tab:Mullang_variance} details the performance of agents across five programming languages: Python, C++, Java, JavaScript, and Go. The primary goal of this track is to assess whether agents can abstract strategic logic independently of implementation syntax.

\begin{enumerate}[leftmargin=10pt]
    \item \textbf{Domain-Specific Language Bias:} We observe a strong correlation between task type and language performance. In competitive tasks, agents achieve peak performance in Python, consistent with the prevalence of Python in AI and game scripting. Conversely, in OIBench, agents implementing algorithms in C++ significantly outperform those using Python. This suggests that agents rely heavily on the statistical distributions of their pre-training data (where competitive programming is dominated by C++) rather than possessing a generalized coding capability.
    \item \textbf{Generalization Gaps:} Significant performance degradation is observed in languages that are less aligned with the specific domain (e.g., JavaScript/Go for games). {Qwen3-Coder} demonstrates the most consistent results across all languages with the lowest average variance, indicating strong cross-language adaptability. In contrast, models such as {GPT-5} and {Doubao-Seed} show considerable fluctuations, reflecting limited generalization ability.
\end{enumerate}

Considering that the core logic for both games and algorithms is inherently language-agnostic, the observed performance gaps indicate a critical failure in algorithmic abstraction. Current agents struggle to decouple strategic reasoning from syntactic implementation, failing to transfer a known solution (e.g., a C++ algorithm) to a different language context (e.g., Python) without performance loss.

\section{Results of ML Track}\label{app:ML}

 \input{tables/MLtrack}

In the ML track, agents are tasked with an end-to-end machine learning pipeline: autonomously generating data, designing architectures, training models in a GPU-enabled environment, and deploying the model for inference. The results are summarized in Table~\ref{tab:MLtrack}. Unlike the Strategy Track where logic is explicit, the ML Track requires agents to manage a complex, multi-stage workflow.

\begin{enumerate}[leftmargin=10pt]
    \item \textbf{Limited Technical Depth.} Despite the availability of GPU resources, we observe that agents generally fail to leverage advanced deep learning techniques. Most agents only managed to implement basic models (e.g., simple MLPs or decision trees) with limited training epochs. Additionally, a subset of agents fail to implement a valid training loop entirely, submitting models with randomized parameters instead of learned weights.
    \item \textbf{Commercial vs. ADK Agent.} A distinct capability gap exists in end-to-end development stability. All commercial agents successfully completed the full lifecycle—from data generation to model deployment. In contrast,  ADK agents struggled with pipeline integration, frequently failing to produce a runnable submission due to environment configuration errors or logical breaks in the workflow.
    \item  \textbf{Performance vs. Rule-Based Track.} Currently, the performance of agents in the ML track significantly lags behind the rule-based strategies observed in the standard Strategy Track. This underperformance is attributed to the agents' inability to autonomously debug and optimize complex ML workflows. Consequently, due to the low baseline quality and high failure rate, we did not conduct multi-round evolutionary experiments for this track. However, as agent capabilities in coding and scientific modeling improve, the ML track will become a critical frontier for evaluating self-evolving intelligence in CATArena.
\end{enumerate}

\input{section/pommerman/pommerman}

\section{Cost and Code Complexity of Participants}\label{app:cost}

We list the agents' cost and code statistics in Table~\ref{tab:model_cost_games_round1} for first round development of standard games, ~\ref{tab:model_cost_games_round2} for second round development of standard games,~\ref{tab:model_cost_games_var_round1} for first round development of variant games and ~\ref{tab:model_cost_games_var_round2} for second round development of variant games. 

We can see that game development token costs show minimal variation, while differences are significant due to model changes. Claude (both minimal and code-based agents) consumes significantly more input tokens than competitors, exceeding the average by over 2 times, while Gemini generates notably more output tokens compared to other models. 
GPT-5 offers the best trade-off between cost and performance. 
Among all agents, second-round game development requires more input tokens, while output token growth remains marginal. In addition, commercial agents consistently use fewer tokens than their minimal-agent counterparts.

In terms of code complexity, agents driven by Claude-4 model consistently surpass other agents in both the number of effective lines of code developed and the time spent considering development strategies. We observe that its development strategies are more sophisticated. Additionally, the complexity of its code increases with each iteration, which indirectly demonstrates the model’s exceptional learning capabilities.

\input{tables/CostRound1}
\input{tables/CostRound2}
\input{tables/CostVar1}
\input{tables/CostVar2}

\section{Full Prompts}\label{app:prompt}
The agent is instructed to develop a competitive game AI based on the provided game environment. The AI must be deployed as an HTTP service with a single-port startup script, follow the official development instructions, and be named with the model prefix. The agent is encouraged to iteratively improve its strategy based on the tournament report and the previous solutions. Full prompt details are in Tables~\ref{tab:game_ai_prompt} for the main leaderboard, ~\ref{tab:ml_game_ai_prompt} for ML track and ~\ref{tab:multila_game_ai_prompt} for multilingual track.

\input{tables/Gameprompt}
\input{tables/MLprompt}
\input{tables/Multilingualprompt}

%% file: tables/evaluationmetric.tex
\begin{table}[h]
\scriptsize 
\caption{Scoring rules of $W$ matrix calculation on six tasks.}
\label{tab:game_scoring}
\centering
\begin{tabular}{|l|p{8cm}|} 
\hline
\multicolumn{1}{|c|}{Environment} & \multicolumn{1}{c|}{Scoring Metric} \\ \hline
Gomoku & \begin{tabular}[c]{@{}l@{}}Pairwise match scoring: \\ Win = 1 point,\; Draw = 0.5 point,\; Lose = 0 point. \end{tabular} \\ \hline
Texas Hold'em & \begin{tabular}[c]{@{}l@{}}Multi-agent batches: \\ score is the average win rate across all tournaments participated.\end{tabular} \\  \hline
Bridge & \begin{tabular}[c]{@{}l@{}}20 VP system: \\ Two opposing pairs’ scores sum to 20, \\ Final score divided by 20, ensuring each pair’s score $\in[0,1]$.\end{tabular} \\ \hline
Chess & \begin{tabular}[c]{@{}l@{}}Pairwise match scoring: \\ Win = 1 point,\; Draw = 0.5 point,\; Lose = 0 point. \end{tabular} \\ \hline
OIBench & \begin{tabular}[c]{@{}l@{}}TimeAUC per problem: CDF of $\log_{10}(t/T_\mathrm{ref})$\\ AUC over $[x_{\min},1]$, normalized by reference AUC. \end{tabular}\\ \hline
SWE-Perf & \begin{tabular}[c]{@{}l@{}}Performance: mean significant speedup across efficiency tests\\ (Mann-Whitney U, $p{<}0.1$) vs baseline. \end{tabular}\\ \hline
\end{tabular}
\end{table}

%% file: tables/TournamentConfig.tex
\begin{table}[h]
\tiny
\caption{Configs of tournament on six tasks.}
\label{tab:game_config}
\centering
\resizebox{0.7\textwidth}{!}{
\begin{tabular}{|l|l|l|}
\hline
\multicolumn{1}{|c|}{Task Type} & \multicolumn{1}{|c|}{Environment} & \multicolumn{1}{c|}{Code Agents}                                                              \\ \hline
Competitive & Gomoku                      & \begin{tabular}[c]{@{}l@{}}Board size: 15$\times$15 \\ Number of pairwise matches: 4 $\times$ 2 \\ Swap black and white pieces after each match\\ Maximum time per move: 10 s\end{tabular} \\ \hline
Competitive & Texas Hold'em               & \begin{tabular}[c]{@{}l@{}}Max players: 12 \\ Rounds: 100 \\ Randomly shuffle seats after each round \\ Initial chips: 2000 \\ Blind increase every 24 hands \\ Max hands per round: 720 or until winner decided \\ Maximum time per move: 3 s\end{tabular} \\ \hline
Competitive & Bridge                       & \begin{tabular}[c]{@{}l@{}}Number of pairwise matches: 12 $\times$ 2 \\ Swap directions of open/closed rooms \\ Use same deck for each pair of matches\\ Maximum time per move: 10 s\end{tabular}       \\ \hline
Competitive & Chess                     & \begin{tabular}[c]{@{}l@{}}Number of pairwise matches: 8 $\times$ 2\\ Swap black and white pieces after each match\\ Maximum moves per game: 200 \\ Maximum time per move: 10 s\end{tabular} \\ \hline
Objective & OIBench & \begin{tabular}[c]{@{}l@{}}Problems per round: 5 \\ Language: C++ \\ Per-test timeout: 2s \\ Score: mean TimeAUC \\ Agent time budget: 60 min \end{tabular} \\ \hline
Objective & SWE-Perf & \begin{tabular}[c]{@{}l@{}}Instance: repo@base\_commit (5 instances per sweep) \\ Efficiency tests from dataset \\ Each test repeated 20$\times$ (+3 warmups) \\ Score: mean significant speedup over baseline
\end{tabular} \\ \hline

\end{tabular}
}
\end{table}

%% file: tables/AgentDescription.tex
\begin{table*}[htb]
\centering
\caption{Agent Specifications and Open-Source Status.}
\label{tab:agentspec}
\resizebox{0.9\textwidth}{!}{%
\begin{tabular}{lllcc}
\toprule
\textbf{Agent Type} & \textbf{Agent Framework} & \textbf{Model} & \textbf{Agent OSS} & \textbf{LLM OSS} \\
\midrule
\multirow{6}{*}{\centering Minimal}  
    & \multirow{6}{*}{basic code tools with ADK framework} 
        & DeepSeek-3.1~\citep{deepseekai2024deepseekv3technicalreport} & \checkmark & \checkmark \\
    &   & Qwen3-Coder-480B~\citep{qwencoder}         &   \checkmark  & \checkmark \\
    &   & Doubao-Seed-1.6~\citep{doubaoseed16}          &  \checkmark   & \ding{55} \\
    &   & GPT-5~\citep{gpt5}                    &  \checkmark   & \ding{55} \\
    &   & Claude-4-Sonnet~\citep{anthropic2025claude4}          &  \checkmark   & \ding{55} \\
    &   & Gemini-2.5-Pro~\citep{gemini25}          &  \checkmark   & \ding{55} \\
\midrule
\multirow{4}{*}{\centering Commercial}    
    & Gemini-CLI~\citep{geminicli} & Gemini-2.5-Pro~\citep{gemini25}     & \checkmark    & \ding{55} \\
    & Claude-Code~\citep{claudecode2025} & Claude-4/3.7 Hybrid~\citep{anthropic2025claude4} & \ding{55} & \ding{55} \\
    & Codex~\citep{codex} & GPT-5~\citep{gpt5}            & \checkmark    & \ding{55} \\
    & Qwen-Code~\citep{qwencode} & Qwen3-Coder-480B~\citep{qwencoder}  & \checkmark    & \checkmark \\
\bottomrule
\end{tabular}
}
\end{table*}

%% file: tables/repeatR1.tex
\begin{table}[h]
\caption{Standard deviation of agent ranking in Round 1 and Round 2 with repeating 4 times.}
\label{tab:std_rounds}

\setlength{\tabcolsep}{5pt} 
\renewcommand{\arraystretch}{1.0} 
\small
\centering
\begin{tabular}{lll|cccccc}
\toprule
\multicolumn{3}{c|}{\textbf{Games}} & \textbf{Gomoku} & \textbf{Hold'em} & \textbf{Bridge} & \textbf{Chess} & \textbf{OIBench} & \textbf{SWE-Perf} \\ 
\midrule

\multicolumn{1}{l|}{\multirow{10}{*}{\rotatebox{90}{Round 1}}}
    & \multirow{6}{*}{\rotatebox{90}{Minimal}}
        & Claude-4-Sonnet & 1.58 & 1.12 & 0.50 & 0.00 & 0.50   & 0.82 \\
    &   & Deepseek-Chat   & 0.83 & 0.83 & 1.22 & 0.00 & 0.50   & 1.83 \\
    &   & Doubao-Seed     & 1.87 & 1.09 & 2.06 & 1.30 & 0.96 & 2.00 \\
    &   & Gemini-2.5-Pro  & 1.50 & 1.12 & 1.12 & 1.22 & 0.58 & 0.96 \\
    &   & GPT-5           & 0.71 & 0.71 & 1.09 & 0.50 & 0.00     & 0.50 \\
    &   & Qwen3-Coder     & 1.48 & 1.22 & 1.09 & 0.83 & 0.50   & 0.82 \\ 
    \cmidrule(lr){2-9}
    & \multirow{4}{*}{\rotatebox{90}{Comm.}}
        & Claude-Code     & 1.12 & 0.43 & 1.79 & 1.73 & 0.96 & - \\
    &   & Codex           & 0.83 & 0.43 & 1.09 & 0.71 & 1.41 & - \\
    &   & Gemini-CLI      & 0.83 & 1.12 & 1.30 & 1.30 & 0.82 & - \\
    &   & Qwen-Coder      & 1.22 & 1.12 & 0.87 & 0.83 & 0.96 & - \\ 
\midrule

\multicolumn{1}{l|}{\multirow{10}{*}{\rotatebox{90}{Round 2}}}
    & \multirow{6}{*}{\rotatebox{90}{Minimal}}
        & Claude-4-Sonnet & 1.48 & 0.83 & 0.50 & 0.43 & 0.50   & 1.23 \\
    &   & Deepseek-Chat   & 1.30 & 0.87 & 1.09 & 0.50 & 0.58 & 0.00 \\
    &   & Doubao-Seed     & 1.09 & 0.87 & 0.71 & 0.50 & 2.16  & 0.00 \\
    &   & Gemini-2.5-Pro  & 1.66 & 1.00 & 2.06 & 0.50 & 0.82 & 1.50 \\
    &   & GPT-5           & 1.00 & 0.43 & 1.64 & 0.50 & 1.00     & 1.16 \\
    &   & Qwen3-Coder     & 1.58 & 0.83 & 0.43 & 0.43 & 0.50   & 0.00 \\ 
    \cmidrule(lr){2-9}
    & \multirow{4}{*}{\rotatebox{90}{Comm.}}
        & Claude-Code     & 1.09 & 1.22 & 1.64 & 0.83 & 0.58 & - \\
    &   & Codex           & 0.43 & 0.50 & 0.83 & 1.09 & 1.91 & - \\
    &   & Gemini-CLI      & 0.83 & 0.83 & 1.41 & 0.43 & 1.73 & - \\
    &   & Qwen-Coder      & 1.66 & 0.43 & 1.00 & 1.41 & 0.58 & - \\ 
\bottomrule

\end{tabular}
\end{table}

%% file: tables/errorRate.tex
\begin{table}[htbp]
\caption{Proportion of HTTP errors encountered during code execution for each agent in Round 1. }
\label{tab:errorRate}
\small
\centering
\begin{tabular}{l|l|cccccccc}
\toprule
\multicolumn{2}{l|}{}                                 & \multicolumn{2}{c}{Gomoku}                                 & \multicolumn{2}{c}{Hold'em}                                & \multicolumn{2}{c}{Bridge}                                 & \multicolumn{2}{c}{Chess}                                  \\
\multicolumn{2}{l|}{\multirow{-2}{*}{HTTP error(\%)}} & \multicolumn{1}{c}{Standard} & \multicolumn{1}{c}{Variant} & \multicolumn{1}{c}{Standard} & \multicolumn{1}{c}{Variant} & \multicolumn{1}{c}{Standard} & \multicolumn{1}{c}{Variant} & \multicolumn{1}{c}{Standard} & \multicolumn{1}{c}{Variant} \\ \midrule
\multicolumn{1}{l|}{\multirow{6}{*}{\rotatebox{90}{{Minimal}}}} & \multicolumn{1}{l|}{Claude-4-Sonnet}                                   & 7                            & 0                           & 0                            & 0                           & 0                            & 0                           & 0                            & 0                           \\
& Deepseek-Chat                                                           & 21                           & 0                           & 0                            & 48                          & 0                            & 0                           & 100                          & 100                         \\
& \multicolumn{1}{l|}{Doubao-Seed}                                        & 17                           & 0                           & 49                           & 93                          & 12                           & 0                           & 0                            & 100                         \\
& Gemini-2.5-Pro                                                                   & 3                            & 100                         & 8                            & 50                          & 0                            & 1                           & 0                            & 0                           \\
& GPT-5                                                                   & 6                            & 0                           & 0                            & 0                           & 0                            & 100                         & 2                            & 21                          \\
& Qwen3-Coder                                                             & 11                           & 0                           & 0                            & 91                          & 2                            & 1                           & 0                            & 0                           \\ \midrule

{\multirow{4}{*}{\rotatebox{90}{{Commercial}}}} & best ADK                                                                      & 0                            & 0                           & 0                            & 0                           &         0                     & 0                           & 0                            & 0                           \\
& Claude-Code                                                                   & 0                            & 0                           & 0                            & 0                           &         0                     & 0                           & 0                            & 0                           \\
& Codex                                                                         & 0                            & 0                           & 0                            & 0                           &        0                      & 0                           & 0                            & 0                           \\
& Gemini-CLI                                                                    & 0                            & 0                           & 0                            & 0                           &      48                        & 0                           & 0                            & 0                           \\
& Qwen-Coder                                                                      & 2                       & 0                           & 0                            & 18                          &         0                     & 17                          & 0                            & 0   \\ \bottomrule                       
\end{tabular}
\end{table}

%% file: tables/variant.tex

\begin{table*}[htb]
\caption{\textbf{Results of variant tasks.} We report the Base Performance ($S_\text{base}$) and Evolutionary Capability ($S_\text{evo}$) of Minimal and Commercial Agents across five variants of standard tasks. Best results in each column are \textbf{bolded}.}
\label{tab:variant}
\centering
\footnotesize  
\resizebox{\textwidth}{!}{%
\setlength{\tabcolsep}{6pt}  
\renewcommand{\arraystretch}{1.0}  
\begin{tabular}{l|l|cc|cc|cc|cc|cc}
\toprule
\multicolumn{2}{c|}{\multirow{2}{*}{\textbf{Agents}}} & \multicolumn{8}{c|}{\textbf{Competitive}} & \multicolumn{2}{c}{\textbf{Objective}} \\
\cmidrule(lr){3-10} \cmidrule(lr){11-12}
\multicolumn{2}{c|}{} & \multicolumn{2}{c|}{\textbf{Gomoku}} & \multicolumn{2}{c|}{\textbf{Hold'em}} & \multicolumn{2}{c|}{\textbf{Bridge}} & \multicolumn{2}{c|}{\textbf{Chess}} & \multicolumn{2}{c}{\textbf{OIBench}} \\
\cmidrule(lr){3-4} \cmidrule(lr){5-6} \cmidrule(lr){7-8} \cmidrule(lr){9-10} \cmidrule(lr){11-12}
\multicolumn{2}{c|}{} & $S_\text{base}$ & $S_\text{evo}$ & $S_\text{base}$ & $S_\text{evo}$ & $S_\text{base}$ & $S_\text{evo}$ & $S_\text{base}$ & $S_\text{evo}$ & $S_\text{base}$ & $S_\text{evo}$ \\
\midrule
\multicolumn{1}{l|}{\multirow{6}{*}{\rotatebox{90}{Minimal}}}
& Claude & \textbf{0.78} & \negval{-0.100} & 0.13 & \posval{+0.017} & \textbf{1.00} & \posval{+0.031} & 0.65 & \negval{-0.014} & 0.23 & \posval{+0.111} \\
& DeepSeek & 0.38 & \posval{+0.063} & 0.00 & \negval{-0.008} & 0.10 & \zeroval{0.000} & 0.10 & \textbf{\posval{+0.020}} & 0.07 & \negval{-0.021} \\
& Doubao & 0.73 & \negval{-0.201} & 0.00 & \zeroval{0.000} & 0.45 & \negval{-0.155} & 0.10 & \posval{+0.018} & 0.22 & \textbf{\posval{+0.181}} \\
& Gemini & 0.00 & \textbf{\posval{+0.159}} & 0.00 & \textbf{\posval{+0.043}} & 0.60 & \negval{-0.011} & \textbf{0.90} & \negval{-0.016} & 0.56 & \negval{-0.028} \\
& GPT & 0.76 & \negval{-0.008} & \textbf{0.87} & \posval{+0.014} & 0.10 & \posval{+0.014} & 0.45 & \posval{+0.004} & \textbf{0.66} & \posval{+0.003} \\
& Qwen & 0.36 & \posval{+0.041} & 0.00 & \negval{-0.008} & 0.76 & \negval{-0.021} & 0.80 & \negval{-0.158} & 0.41 & \negval{-0.011} \\
\midrule
\multirow{5}{*}{\rotatebox{90}{Commercial}} 
& Best ADK Agent & \textbf{0.75} & \posval{+0.012} & \textbf{0.46} & \posval{+0.010} & 0.00 & \posval{+0.109} & \textbf{1.00} & \negval{-0.182} & \textbf{0.75} & \posval{+0.036}\\
& Claude Code &  0.66 &\posval{+0.093} & 0.00 & \posval{+0.038} & \textbf{0.93} & \posval{+0.024} & 0.44 & \negval{-0.058} & 0.74 & \posval{+0.019}  \\
& Codex &  0.69 &\negval{-0.087} & 0.17 & \posval{+0.030} & 0.50 & \posval{+0.013} & 0.34 & \posval{+0.029} & 0.00 & \zeroval{0.000} \\
& Gemini CLI & 0.19 &\negval{-0.058} &  0.37 & \zeroval{0.000} & 0.83 & \negval{-0.191}& 0.38 & \negval{-0.012} & 0.66 & \textbf{\posval{+0.160}} \\
& Qwen Code &  0.22 &\textbf{\posval{+0.146}} & 0.00 & \textbf{\posval{+0.049}} & 0.25 & \textbf{\posval{+0.144}} & 0.34 & \textbf{\posval{+0.011}} & 0.70 & \negval{-0.042}  \\

\bottomrule
\end{tabular}
}
\end{table*}

%% file: tables/MulLang.tex
\begin{table}[h]
  \centering
  \caption{Scores of agents on games across programming languages, with variance analysis.}
  \label{tab:Mullang_variance}
  \setlength{\tabcolsep}{4pt}
  \renewcommand{\arraystretch}{1.1}
  \resizebox{0.99\textwidth}{!}{
    \begin{tabular}{l|l|cccc|cccc|cccc|c}
      \toprule
      \multicolumn{2}{c|}{\textbf{Agent}} 
      & \multicolumn{4}{c|}{\textbf{Gomoku}}
      & \multicolumn{4}{c|}{\textbf{Hold'em}}
      & \multicolumn{4}{c|}{\textbf{Bridge}}
      & \textbf{Avg. Variance}$\downarrow$ \\
      \cmidrule(lr){3-6} \cmidrule(lr){7-10} \cmidrule(lr){11-14}
      & & Python$\uparrow$ & JS$\uparrow$ & Go$\uparrow$ & Var.$\downarrow$
        & Python$\uparrow$ & JS$\uparrow$ & Go$\uparrow$ & Var.$\downarrow$
        & Python$\uparrow$ & JS$\uparrow$ & Go$\uparrow$ & Var.$\downarrow$
        & \\
      \midrule
      \multicolumn{1}{l|}{\multirow{6}{*}{\rotatebox{90}{{Minimal}}}}
        & Claude-4-Sonnet     & 1.000 & 0.250 & 0.250 & 0.125 & 0.360 & 0.640 & 0.000 & 0.069 & 1.000 & 0.250 & 0.250 & 0.125 & 0.106 \\
        & DeepSeek-Chat       & 1.000 & 0.250 & 0.250 & 0.125 & 1.000 & 0.000 & 0.000 & 0.222 & 0.500 & 0.500 & 0.500 & 0.000 & 0.116 \\
        & Doubao-Seed         & 1.000 & 0.500 & 0.000 & 0.167 & 1.000 & 0.000 & 0.000 & 0.222 & 0.500 & 1.000 & 0.000 & 0.167 & 0.185 \\
        & Gemini-2.5-Pro      & 0.750 & 0.750 & 0.000 & 0.125 & 0.010 & 0.990 & 0.000 & 0.216 & 1.000 & 0.250 & 0.250 & 0.125 & 0.155 \\
        & GPT-5               & 0.500 & 0.000 & 1.000 & 0.167 & 1.000 & 0.000 & 0.000 & 0.222 & 1.000 & 0.500 & 0.000 & 0.167 & 0.185 \\
        & Qwen3-Coder         & 1.000 & 0.250 & 0.250 & 0.125 & 0.610 & 0.290 & 0.100 & 0.044 & 0.688 & 0.000 & 0.812 & 0.128 & \textbf{0.099} \\
      \midrule \midrule
      \multicolumn{1}{l|}{\multirow{4}{*}{\rotatebox{90}{{Commercial}}}}
        & Claude-Code         & 1.000 & 0.250 & 0.250 & 0.125 & 0.200 & 0.020 & 0.780 & 0.105 & 1.000 & 0.250 & 0.250 & 0.125 & 0.118 \\
        & Codex               & 0.000 & 0.500 & 1.000 & 0.167 & 0.200 & 0.030 & 0.770 & 0.100 & 1.000 & 0.000 & 0.500 & 0.167 & 0.145 \\
        & Gemini-CLI          & 0.750 & 0.750 & 0.000 & 0.125 & 1.000 & 0.000 & 0.000 & 0.222 & 1.000 & 0.250 & 0.250 & 0.125 & 0.157 \\
        & Qwen-Coder           & 1.000 & 0.250 & 0.250 & 0.125 & 1.000 & 0.000 & 0.000 & 0.222 & 0.975 & 0.000 & 0.525 & 0.159 & 0.169 \\
      \bottomrule
    \end{tabular}
  }
\end{table}

%% file: tables/MLtrack.tex
\begin{table}[htb]
  \centering
  \caption{ML ability scores and average rankings of agents.}
  \label{tab:MLtrack}
  \setlength{\tabcolsep}{5pt}
  \renewcommand{\arraystretch}{1.0}
  \resizebox{0.6\textwidth}{!}{
    \begin{tabular}{l|l|cccc|c}
      \toprule
      \multicolumn{2}{c|}{\textbf{Agent}} & \textbf{Gomoku}$\uparrow$ & \textbf{Hold'em}$\uparrow$ & \textbf{Bridge}$\uparrow$ & 
      \textbf{Chess}$\uparrow$ & \textbf{Avg. Ranking}$\downarrow$ \\ 
      \midrule
      \multicolumn{1}{l|}{\multirow{6}{*}{\rotatebox{90}{\textbf{Minimal}}}}
          & Claude-4-Sonnet & \textbf{0.787} & 0.360 & 0.600 & 0.700 & 2.25 \\
          & DeepSeek-Chat   & 0.612 & 0.000 & 0.170 & 0.100 & 4.25 \\
          & Doubao-Seed     & 0.375 & 0.110 & 0.290 & 0.100 & 4.25 \\
          & Gemini-2.5-Pro  & 0.000 & 0.000 & 0.140 & 0.675 & 5.00 \\
          & GPT-5           & 0.625 & \textbf{0.530} & \textbf{0.900} & 0.700 & \textbf{1.50} \\
          & Qwen3-Coder     & 0.600 & 0.000 & \textbf{0.900} & \textbf{0.725} & 2.50 \\ 
      \midrule \midrule
      \multicolumn{1}{l|}{\multirow{5}{*}{\rotatebox{90}{\textbf{Commercial}}}}
          & best ADK        & \textbf{0.750} & 0.190 & \textbf{0.700} & \textbf{0.656} & \textbf{1.25} \\
          & Claude-Code     & 0.578 & 0.170 & 0.000 & 0.406 & 4.00 \\
          & Codex           & 0.484 & 0.190 & 0.400 & 0.469 & 3.00 \\
          & Gemini-CLI      & 0.187 & \textbf{0.280} & 0.200 & 0.438 & 3.50 \\
          & Qwen-Coder      & 0.500 & 0.170 & \textbf{0.700} & 0.531 & 2.50 \\
      \bottomrule
    \end{tabular}
  }
\end{table}

%% file: section/pommerman/pommerman.tex
\section{Agent performance on Hard Games: Pommerman}\label{app:pommerman}

\begin{figure}[htbp]
    \centering
    \begin{subfigure}[b]{0.45\textwidth}
        \centering
        \includegraphics[width=\linewidth]{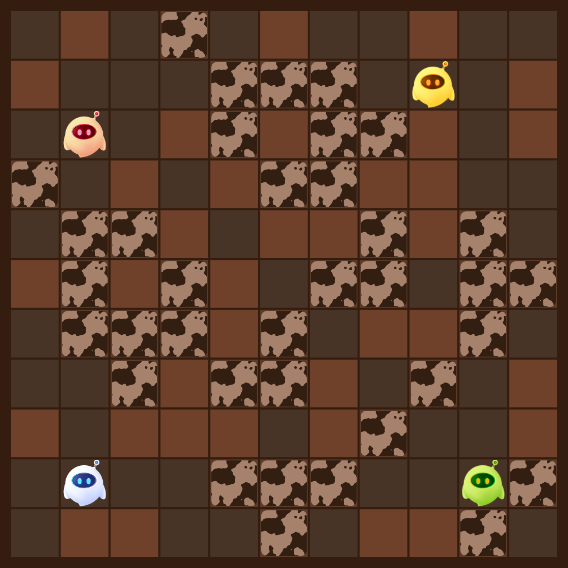}
        \caption{Start Status~(step=0)}
        \label{fig:start}
    \end{subfigure}
    \hfill
    \begin{subfigure}[b]{0.45\textwidth}
        \centering
        \includegraphics[width=\linewidth]{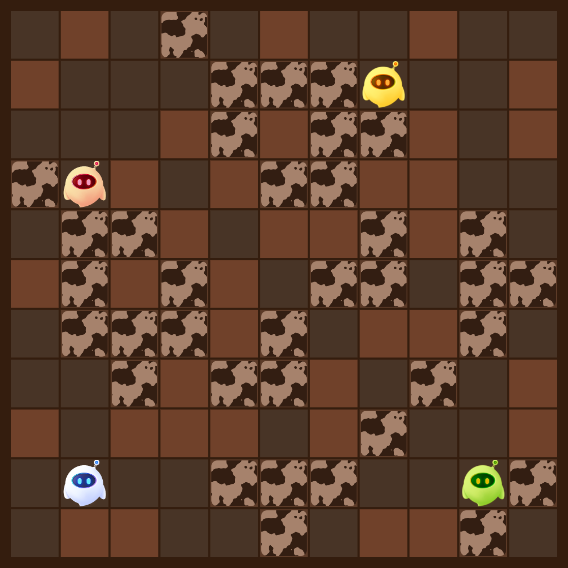}
        \caption{End Status~(step=800)}
        \label{fig:end}
    \end{subfigure}
    \caption{One Pommerman round demo. \textbf{All agents failed to generate functional strategies}, resulting in aimless movement or inactivity.}
    \label{fig:game_screens}
\end{figure}

Beyond the fundamental board games discussed in previous sections, we further investigate more sophisticated competitive environments that necessitate the ability to perceive complex scenarios and synthesize highly strategic code. Specifically, we utilize the \textbf{Pommerman} environment~\cite{DBLP:journals/corr/abs-1809-07124}, which was originally designed for multi-agent AI research and featured in the NeurIPS 2018 Competition Track\footnote{\url{https://neurips.cc/Conferences/2018/CompetitionTrack}}. Inspired by the classic Nintendo game \textit{Bomberman}, Pommerman is played on a randomly generated symmetric $11 \times 11$ grid. The environment contains both rigid walls (indestructible and impassable) and wooden walls (destructible by bombs). Each agent can execute one of six actions per turn: \textit{Stop, Up, Left, Down, Right,} and \textit{Bomb}, with the ultimate objective of being the sole survivor.

\paragraph{Zero-Shot Performance Failure $S_\text{base}$} 
The environment's infrastructure comprises approximately 4,500 lines of Python code, while a proficient human-authored solution typically requires around 1,000 lines of strategic logic to achieve plausible gameplay. We task the agents with developing autonomous AI strategies for Pommerman; however, almost all agent-generated strategies proved ineffective during live evaluation. Our observations indicate that these results typically fall into three failure paradigms: 

\begin{itemize}
    \item \textbf{Random Walk:} The agent moves aimlessly across the field without ever deploying bombs.
    \item \textbf{Self-Destruction:} The agent successfully deploys bombs but fails to execute evasion maneuvers, resulting in self-elimination.
    \item \textbf{Inactivity:} The agent fails to produce valid operations within the turn limit, remaining stationary throughout the match.
\end{itemize}

\paragraph{Code Analysis and Implications}
The generated code ranged from 100 to 400 lines, relying mostly on random actions or rudimentary Breadth-First Search (BFS). While {Claude} attempted to implement modular functions for tactical value estimation, its code was riddled with logical bugs that led to runtime errors or inactivity.

These results highlight a critical cold start problem in complex environments: current LLMs cannot bridge the gap between understanding game rules and implementing a minimum viable strategy (MVP).Without a functional baseline, the evolutionary feedback loop cannot be initiated, validating our decision to exclude such high-complexity tasks from the main evolutionary experiments.

Figure~\ref{fig:game_screens} illustrates a typical match where agents fail to engage meaningfully.
The blue(Claude's) and green(OpenAI's) pommerman stay still, the red(Gemini's) one find a more valuable place and then become inactive, the yellow(Qwen's) pommerman walk in random.

%% file: tables/CostRound1.tex
\begin{sidewaystable}[htbp]
\centering
\caption{Model cost statistics on standard tasks in Round 1.}
\label{tab:model_cost_games_round1}
\scriptsize
\begin{tabular}{lcccccccccc}
\hline
\textbf{Metric} & \textbf{Claude-4-Sonnet} & \textbf{Deepseek-Chat} & \textbf{Doubao-Seed} & \textbf{Gemini-2.5-Pro} & \textbf{GPT-5} & \textbf{Qwen3-Coder} & \textbf{Claude-Code} & \textbf{Codex} & \textbf{Gemini-CLI} & \textbf{Qwen-Coder} \\
\hline
\multicolumn{11}{c}{\textbf{Gomoku}} \\
Total input tokens & 825964 & 34266 & 83307 & 1074278 & 392378 & 251590 & 925166 & - & 169567 & 623393 \\
Session tokens     & 31571 & 8256 & 20186 & 27481 & 25201 & 20013 & 29340 & - & 21590 & 26478 \\
Total output tokens& 13777 & 1400 & 9432 & 25952 & 14184 & 6338 & 11746 & - & 2303 & 8002 \\
Total time (s)     & 441.916 & 29.021 & 206.159 & 858.027 & 280.399 & 135.770 & 323.200 & - & 519.100 & 607.200 \\
Tools used         & 68 & 12 & 20 & 118 & 44 & 32 & 38 & 24 & 8 & 26 \\
Valid lines of code& 574 & 69 & 405 & 65 & 421 & 425 & 350 & 347 & 101 & 335 \\
Avg thinking time (s) & 0.0337 & 0.0026 & 2.9848 & 0.0026 & 1.9681 & 0.0903 & 0.0053 & 0.7026 & 0.1151 & 0.0036 \\
\hline
\multicolumn{11}{c}{\textbf{Texas Hold'em Poker}} \\
Total input tokens & 966609.5 & 44883 & 97274 & 168565 & 345444 & 326508 & 296329.5 & - & 840690.5 & 1268906.5 \\
Session tokens     & 45384 & 9388 & 14413 & 14073 & 26418 & 16688 & 29663.5 & - & 27828 & 23646.5 \\
Total output tokens& 20974 & 985 & 10370 & 40762 & 24178 & 4472 & 8537 & - & 11833.5 & 12772 \\
Total time (s)     & 489.150 & 25.385 & 256.145 & 481.315 & 473.910 & 131.630 & 143.420 & - & 365.485 & 474.075 \\
Tools used         & 38 & 8 & 12 & 15 & 20 & 24 & 12 & 19 & 39 & 48 \\
Valid lines of code& 742 & 21 & 169 & 92 & 346 & 164 & 122 & 318 & 300 & 220 \\
Avg thinking time (s) & 0.0026 & 0.0023 & 0.0030 & 0.0025 & 0.1004 & 0.0025 & 0.0028 & 0.0021 & 0.0024 & 0.0023 \\
\hline
\multicolumn{11}{c}{\textbf{Bridge}} \\
Total input tokens & 1429350 & 18027 & 81206 & 266506 & 258633 & 114679 & 618991 & - & 204913 & 71696 \\
Session tokens     & 38087 & 15077 & 24374 & 29486 & 35448 & 20358 & 34263 & - & 46945 & 34937 \\
Total output tokens& 17039 & 1131 & 19247 & 12713 & 3954 & 5394 & 12284 & - & 13222 & 20308 \\
Total time (s)     & 727.550 & 3.523 & 333.361 & 191.775 & 128.895 & 334.394 & 274.200 & - & 188.700 & 613.000 \\
Tools used         & 94 & 6 & 14 & 32 & 26 & 16 & 25 & 24 & 5 & 32 \\
Valid lines of code& 521 & 0 & 552 & 280 & 245 & 271 & 464 & 362 & 349 & 549 \\
Avg thinking time (s) & 0.0095 & 0.0000 & 0.0105 & 0.0094 & 0.0095 & 0.0093 & 0.0091 & 0.0091 & 0.0093 & 0.0093 \\
\hline
\multicolumn{11}{c}{\textbf{Chess}} \\
Total input tokens & 1612835 & 49655 & 86674 & 346890 & 484611 & 267156 & 709737 & - & 153825 & 23489 \\
Session tokens     & 38729 & 9953 & 14799 & 93292 & 39962 & 20781 & 29818 & - & 25418 & 23620 \\
Total output tokens& 16584 & 1257 & 14411 & 80997 & 28695 & 6624 & 13017 & - & 4727 & 6845 \\
Total time (s)     & 617.339 & 26.882 & 287.976 & 356.551 & 444.932 & 141.417 & 285.593 & - & 162.342 & 142.344 \\
Tools used         & 110 & 14 & 16 & 40 & 44 & 34 & 32 & 14 & 6 & 18 \\
Valid lines of code& 734 & 0 & 292 & 359 & 437 & 360 & 393 & 283 & 262 & 325 \\
Avg thinking time (s) & 1.3500 & 0.0000 & 0.0030 & 0.0030 & 0.2690 & 0.0020 & 1.9730 & 0.0030 & 0.0030 & 0.0050 \\
\hline
\multicolumn{11}{c}{\textbf{OIBench}} \\
Total input tokens & 1101801 & 127528 & 1056516 & 8320878 & 193403 & 472132 & 1591886 & - & - & - \\
Session tokens     & 1117511 & 130209 & 1135372 & 8387950 & 216516 & 479643 & 1603116 & - & - & - \\
Total output tokens& 15710 & 2681 & 78856 & 67072 & 23113 & 7511 & 11230 & - & - & - \\
Total time (s)     & 663.28 & 144.24 & 1982.12 & 3603.76 & 659.12 & 346.27 & 529.58 & - & - & - \\
Tools used         & 39 & 19 & 22 & 95 & 17 & 25 & 40 & - & - & - \\
Valid lines of code& 312 & 149 & 242 & 375 & 167 & 273 & 381 & - & 548 & 135 \\
Avg thinking time (s) & - & - & - & - & - & - & - & - & - & - \\
\hline
\multicolumn{11}{c}{\textbf{SWE-Perf}} \\
Total input tokens & 12566767 & 1233619 & 2867580 & 1109026 & 5945407 & 12246895 & - & - & - & - \\
Session tokens     & 12622336 & 1240596 & 2977490 & 1155080 & 6049146 & 12308897 & - & - & - & - \\
Total output tokens& 55569 & 6977 & 109910 & 46054 & 103739 & 62002 & - & - & - & - \\
Total time (s)     & 5028.39 & 2187.02 & 3284.32 & 2661.64 & 4082.8 & 4318.23 & - & - & - & - \\
Tools used         & 123 & 23 & 31 & 14 & 76 & 138 & - & - & - & - \\
Valid lines of code& 257 & 0 & 31 & 100 & 125 & 133 & - & - & - & - \\
Avg thinking time (s) & - & - & - & - & - & - & - & - & - & - \\
\hline
\end{tabular}
\end{sidewaystable}

%% file: tables/CostRound2.tex
\begin{sidewaystable}[htbp]
\centering
\caption{Model cost statistics on standard tasks in Round 2.}
\label{tab:model_cost_games_round2}
\scriptsize
\begin{tabular}{lcccccccccc}
\hline
\textbf{Metric} & \textbf{Claude-4-Sonnet} & \textbf{Deepseek-Chat} & \textbf{Doubao-Seed} & \textbf{Gemini-2.5-Pro} & \textbf{GPT-5} & \textbf{Qwen3-Coder} & \textbf{Claude-Code} & \textbf{Codex} & \textbf{Gemini-CLI} & \textbf{Qwen-Coder} \\
\hline
\multicolumn{11}{c}{\textbf{Gomoku}} \\
Total input tokens & 1924785 & 170294 & 625384 & 589934 & 1351710 & 4859091 & 761468 & - & 514517 & 573759 \\
Session tokens     & 102767 & 3495 & 84577 & 38760 & 78092 & 151664 & 37085 & - & 53402 & 27637 \\
Total output tokens& 14154 & 1608 & 16828 & 17098 & 10126 & 8230 & 11705 & - & 5180 & 7006 \\
Total time (s)     & 433.112 & 765.076 & 436.142 & 1195.109 & 387.641 & 918.349 & 265.7 & - & 104.4 & 273.5 \\
Tools used         & 58 & 62 & 32 & 58 & 48 & 82 & 30 & 22 & 12 & 23 \\
Valid lines of code& 617 & 69 & 527 & 305 & 467 & 368 & 493 & 302 & 251 & 376 \\
Avg thinking time (s) & 1.9630 & 0.0026 & 2.1291 & 0.0000 & 1.5888 & 0.3499 & 0.7542 & 0.2172 & 0.0080 & 0.0047 \\
\hline
\multicolumn{11}{c}{\textbf{Texas Hold'em Poker}} \\
Total input tokens & 1497923 & 56126 & 225806 & 2076630 & 1328520 & 3671952 & 1143571 & - & 1048050 & 1071708 \\
Session tokens     & 143575 & 9819 & 57890 & 240057 & 113649 & 323278 & 53929 & - & 40566 & 49161 \\
Total output tokens& 7309 & 1151.5 & 10263.5 & 46280.5 & 13269.5 & 5986.5 & 7349.5 & - & 12617 & 6954 \\
Total time (s)     & 308.05 & 45.935 & 327.225 & 1171.58 & 691.74 & 1057.84 & 866.715 & - & 1453.08 & 163.275 \\
Tools used         & 16 & 11 & 10 & 17 & 21 & 25 & 28 & 18 & 41 & 27 \\
Valid lines of code& 283 & 21 & 110 & 156 & 362 & 293 & 264 & 319 & 363 & 315 \\
Avg thinking time (s) & 0.0026 & 0.0024 & 0.0024 & 0.0019 & 0.0912 & 0.0026 & 0.0023 & 0.0024 & 0.0025 & 0.0025 \\
\hline
\multicolumn{11}{c}{\textbf{Bridge}} \\
Total input tokens & 740733 & 18388 & 184814 & 277299 & 900865 & 720015 & 940709 & - & 275340 & 105056 \\
Session tokens     & 28130 & 15318 & 4076 & 131089 & 119593 & 62734 & 38388 & - & 49946 & 63756 \\
Total output tokens& 14200 & 1178 & 18169 & 7520 & 4983 & 10270 & 15530 & - & 8161 & 25125 \\
Total time (s)     & 397.164 & 6.097 & 391.047 & 162.147 & 367.204 & 1863.073 & 365.9 & - & 169.0 & 1337.7 \\
Tools used         & 48 & 8 & 28 & 26 & 68 & 44 & 33 & 17 & 6 & 38 \\
Valid lines of code& 650 & 0 & 731 & 271 & 245 & 288 & 700 & 367 & 280 & 1112 \\
Avg thinking time (s) & 0.0101 & 0.0000 & 0.0135 & 0.0099 & 0.0096 & 0.0099 & 0.0105 & 0.0105 & 0.0102 & 0.0106 \\
\hline
\multicolumn{11}{c}{\textbf{Chess}} \\
Total input tokens & 2118018 & 45051 & 154263 & 426659 & 390211 & 1014853 & 4129953 & - & 308116 & 31752 \\
Session tokens     & 101195 & 3954 & 28503 & 26538 & 35482 & 36705 & 72658 & - & 33919 & 31877 \\
Total output tokens& 13067 & 412 & 10063 & 13435 & 16798 & 11018 & 25438 & - & 6283 & 7250 \\
Total time (s)     & 1352.194 & 522.132 & 230.664 & 931.231 & 214.792 & 263.665 & 906.521 & - & 420.146 & 177.453 \\
Tools used         & 44 & 26 & 28 & 44 & 34 & 80 & 78 & 14 & 13 & 28 \\
Valid lines of code& 647 & 0 & 305 & 347 & 559 & 499 & 736 & 351 & 299 & 335 \\
Avg thinking time (s) & 0.8240 & 0.0000 & 0.0020 & 0.0000 & 0.0030 & 0.0020 & 0.0030 & 0.0040 & 1.8470 & 0.0050 \\
\hline
\multicolumn{11}{c}{\textbf{OIBench}} \\
Total input tokens & 689268 & 56536 & 734892 & 445268 & 331248 & 434818 & 1667173 & - & - & - \\
Session tokens     & 697796 & 57894 & 817803 & 491666 & 372662 & 441152 & 1674074 & - & - & - \\
Total output tokens& 8528 & 1358 & 82911 & 46398 & 41414 & 6334 & 6901 & - & - & - \\
Total time (s)     & 499.51 & 68.95 & 2024.68 & 1868.17 & 1204.64 & 249.76 & 471.25 & - & - & - \\
Tools used         & 29 & 14 & 34 & 23 & 22 & 25 & 46 & - & - & - \\
Valid lines of code& 278 & 0 & 398 & 271 & 339 & 306 & 419 & - & 780 & 420 \\
Avg thinking time (s) & - & - & - & - & - & - & - & - & - & - \\
\hline
\multicolumn{11}{c}{\textbf{SWE-Perf}} \\
Total input tokens & 6720993 & 633967 & 1735119 & 435355 & 4530952 & 16977184 & - & - & - & - \\
Session tokens     & 6738614 & 635214 & 1803480 & 506936 & 4593913 & 17070695 & - & - & - & - \\
Total output tokens& 17621 & 1247 & 68361 & 71581 & 62961 & 93511 & - & - & - & - \\
Total time (s)     & 3939.86 & 263.16 & 2344.74 & 2377.89 & 3652.3 & 2546.19 & - & - & - & - \\
Tools used         & 79 & 15 & 18 & 17 & 48 & 156 & - & - & - & - \\
Valid lines of code& 34 & 0 & 13 & 39 & 31 & 62 & - & - & - & - \\
Avg thinking time (s) & - & - & - & - & - & - & - & - & - & - \\
\hline

\end{tabular}
\end{sidewaystable}

%% file: tables/CostVar1.tex
\begin{sidewaystable}[htbp]
\centering
\caption{Model cost statistics on variant tasks in Round 1.}
\label{tab:model_cost_games_var_round1}
\scriptsize
\begin{tabular}{lcccccccccc}
\hline
\textbf{Metric} & \textbf{Claude-4-Sonnet} & \textbf{Deepseek-Chat} & \textbf{Doubao-Seed} & \textbf{Gemini-2.5-Pro} & \textbf{GPT-5} & \textbf{Qwen3-Coder} & \textbf{Claude-Code} & \textbf{Codex} & \textbf{Gemini-CLI} & \textbf{Qwen-Coder} \\
\hline
\multicolumn{11}{c}{\textbf{Gomoku}} \\
Total input tokens & 833531 & 35489 & 83792 & 342027 & 539388 & 68087 & 495982 & - & 256919 & 23991 \\
Complete tokens     & 22521 & 8587 & 23512 & 27175 & 27291 & 11468 & 27275 & - & 39232 & 24082 \\
Total output tokens & 10900 & 1394 & 11740 & 13553 & 19676 & 1658 & 8153 & - & 3522 & 7436 \\
Total time (s)      & 307.319 & 24.007 & 255.817 & 809.100 & 365.077 & 73.260 & 191.800 & - & 74.300 & 140.200 \\
Tools used          & 64 & 14 & 20 & 36 & 52 & 14 & 23 & 13 & 7 & 18 \\
Valid lines of code  & 859 & 64 & 746 & 414 & 301 & 298 & 254 & 313 & 156 & 431 \\
Avg thinking time (s)& 0.0473 & 0.0026 & 0.0333 & 6.7802 & 0.0904 & 0.0048 & 0.0106 & 0.0051 & 0.0124 & 0.7938 \\
\hline
\multicolumn{11}{c}{\textbf{Texas Hold'em Poker}} \\
Total input tokens & 1182600 & 39782 & 89270 & 2582120 & 191064 & 385246 & 160410 & - & 864568 & 775880 \\
Complete tokens     & 39575 & 9174 & 14448 & 32547 & 16530 & 16013 & 24287 & - & 25654 & 25308 \\
Total output tokens & 13110 & 962 & 9897 & 52457 & 16407 & 4361 & 5801 & - & 10699 & 7568 \\
Total time (s)      & 398.130 & 20.195 & 217.600 & 1337.345 & 558.055 & 124.640 & 84.920 & - & 319.745 & 270.315 \\
Tools used          & 42 & 7 & 11 & 86 & 17 & 30 & 7 & 19 & 42 & 34 \\
Valid lines of code  & 469 & 21 & 204 & 98 & 302 & 144 & 167 & 256 & 232 & 243 \\
Avg thinking time (s)& 0.0026 & 0.0022 & 0.0011 & 0.0486 & 0.0019 & 0.0011 & 0.0025 & 0.0801 & 0.0028 & 0.0021 \\
\hline
\multicolumn{11}{c}{\textbf{Bridge}} \\
Total input tokens & 475377 & 16805 & 122427 & 232284 & 110768 & 653245 & 1908005 & - & 173628 & 26896 \\
Complete tokens     & 37640 & 13764 & 36846 & 22593 & 4433 & 4438 & 53465 & - & 33184 & 27023 \\
Total output tokens & 16704 & 1137 & 7583 & 10020 & 2378 & 22823 & 18489 & - & 7412 & 5191 \\
Total time (s)      & 298.800 & 7.721 & 158.254 & 483.948 & 304.010 & 7111.237 & 462.300 & - & 229.200 & 129.600 \\
Tools used          & 34 & 6 & 16 & 32 & 40 & 308 & 55 & 16 & 5 & 22 \\
Valid lines of code  & 627 & 0 & 291 & 213 & 133 & 165 & 469 & 368 & 212 & 224 \\
Avg thinking time (s)& 0.0071 & 0.0000 & 0.0068 & 0.0068 & 0.0096 & 0.0067 & 0.0089 & 0.0089 & 0.0060 & 0.0092 \\
\hline
\multicolumn{11}{c}{\textbf{Chess}} \\
Total input tokens & 1000994 & 38977 & 84149 & 432950 & 229279 & 253674 & 979611 & - & 184178 & 115022 \\
Complete tokens     & 38485 & 7761 & 17321 & 43002 & 20520 & 17592 & 34597 & - & 26984 & 19923 \\
Total output tokens & 16355 & 1125 & 10518 & 32332 & 16499 & 5948 & 13010 & - & 6014 & 8032 \\
Total time (s)      & 473.286 & 23.186 & 241.221 & 729.121 & 283.023 & 144.435 & 357.587 & - & 139.114 & 368.952 \\
Tools used          & 74 & 14 & 14 & 40 & 32 & 38 & 41 & 12 & 8 & 19 \\
Valid lines of code  & 719 & 41 & 399 & 398 & 475 & 343 & 410 & 427 & 152 & 392 \\
Avg thinking time (s)& 0.7520 & 0.0000 & 0.0000 & 0.2560 & 1.1190 & 0.0020 & 1.9660 & 0.0040 & 0.0040 & 0.0050 \\
\hline
\multicolumn{11}{c}{\textbf{OIBench}} \\
Total input tokens & 1360571 & 33086 & 7669553 & 339500 & 325721 & 721710 & 2035052 & 27754 & - & - \\
Complete tokens & 1380136 & 35679 & 7786790 & 406925 & 373777 & 732945 & 2052561 & 28256 & - & - \\
Total output tokens & 19565 & 2593 & 117237 & 67425 & 48056 & 11235 & 17509 & 502 & - & - \\
Total time (s) & 2433.550 & 80.090 & 3607.680 & 3600.980 & 1344.090 & 834.490 & 541.540 & 56.040 & - & - \\
Tools used & 44 & 8 & 127 & 27 & 23 & 34 & 46 & 7 & - & - \\
Valid lines of code & 224 & 82 & 421 & 248 & 276 & 271 & 248 & 0 & 405 & 371 \\
Avg thinking time (s) & - & - & - & - & - & - & - & - & - & - \\
\hline
\end{tabular}
\end{sidewaystable}

%% file: tables/CostVar2.tex
\begin{sidewaystable}[htbp]
\centering
\caption{Model cost statistics on variant tasks in Round 2.}
\label{tab:model_cost_games_var_round2}
\scriptsize
\begin{tabular}{lcccccccccc}
\hline
\textbf{Metric} & \textbf{Claude-4-Sonnet} & \textbf{Deepseek-Chat} & \textbf{Doubao-Seed} & \textbf{Gemini-2.5-Pro} & \textbf{GPT-5} & \textbf{Qwen3-Coder} & \textbf{Claude-Code} & \textbf{Codex} & \textbf{Gemini-CLI} & \textbf{Qwen-Coder} \\
\hline
\multicolumn{11}{c}{\textbf{Gomoku}} \\
Total input tokens & 1298765 & 111940 & 680676 & 545504 & 606704 & 1542046 & 805252 & - & 227722 & 873934 \\
Complete tokens    & 132219 & 11100 & 121218 & 24003 & 82812 & 4309 & 38629 & - & 41093 & 32599 \\
Total output tokens & 7118 & 1902 & 26565 & 18241 & 4189 & 22014 & 11529 & - & 5554 & 8531 \\
Total time (s)      & 261.323 & 76.973 & 538.087 & 325.892 & 179.801 & 25787.922 & 294.400 & - & 80.200 & 836.300 \\
Tools used          & 46 & 30 & 28 & 70 & 38 & 538 & 30 & 23 & 6 & 30 \\
Valid lines of code  & 264 & 64 & 535 & 111 & 276 & 43 & 328 & 302 & 210 & 384 \\
Avg thinking time (s)& 0.0212 & 0.0030 & 0.0855 & 0.0000 & 0.0904 & 0.0000 & 0.0884 & 0.0162 & 0.0070 & 0.0071 \\
\hline
\multicolumn{11}{c}{\textbf{Texas Hold'em Poker}} \\
Total input tokens & 1627500 & 50339 & 656196 & 675580 & 1452101 & 1822342 & 448924 & - & 1725678 & 2472353 \\
Complete tokens    & 143514 & 11897 & 115933 & 103332 & 111407 & 118321 & 48631 & - & 48378 & 67453 \\
Total output tokens & 8180 & 1130 & 11609 & 18771 & 12897 & 6227 & 5880 & - & 16033 & 10296 \\
Total time (s)      & 367.720 & 27.960 & 359.490 & 515.000 & 408.505 & 294.910 & 612.525 & - & 670.535 & 283.440 \\
Tools used          & 18 & 10 & 16 & 20 & 22 & 27 & 16 & 23 & 53 & 48 \\
Valid lines of code  & 289 & 22 & 272 & 105 & 324 & 295 & 167 & 256 & 232 & 243 \\
Avg thinking time (s)& 0.0025 & 0.0019 & 0.0019 & 0.0031 & 0.0019 & 0.0026 & 0.0025 & 0.0801 & 0.0028 & 0.0021 \\
\hline
\multicolumn{11}{c}{\textbf{Bridge}} \\
Total input tokens & 706814 & 17186 & 32819 & 775688 & 433901 & 866917 & 1110504 & - & 189659 & 48922 \\
Complete tokens    & 41164 & 14021 & 14545 & 80161 & 4141 & 4944 & 45001 & - & 46073 & 49034 \\
Total output tokens & 15867 & 1188 & 13212 & 39919 & 1485 & 13928 & 17233 & - & 7966 & 7252 \\
Total time (s)      & 359.864 & 6.074 & 189.645 & 279.231 & 1106.275 & 3238.660 & 411.500 & - & 77.700 & 220.000 \\
Tools used          & 52 & 8 & 12 & 46 & 112 & 158 & 38 & 21 & 4 & 25 \\
Valid lines of code  & 603 & 0 & 233 & 396 & 166 & 218 & 747 & 295 & 441 & 531 \\
Avg thinking time (s)& 0.0069 & 0.0000 & 0.0112 & 0.0067 & 0.0065 & 0.0101 & 0.0093 & 0.0091 & 0.0091 & 0.0092 \\
\hline
\multicolumn{11}{c}{\textbf{Chess}} \\
Total input tokens & 1475420 & 34542 & 172443 & 400275 & 473325 & 1682584 & 2213785 & - & 247641 & 109313 \\
Complete tokens    & 56660 & 8283 & 31582 & 97638 & 32082 & 44129 & 43227 & - & 23389 & 8140 \\
Total output tokens & 22989 & 1473 & 16610 & 35371 & 16469 & 14241 & 24090 & - & 4834 & 9129 \\
Total time (s)      & 696.717 & 32.412 & 366.723 & 1093.706 & 350.225 & 511.943 & 605.236 & - & 237.026 & 427.591 \\
Tools used          & 82 & 14 & 30 & 62 & 44 & 116 & 67 & 16 & 13 & 29 \\
Valid lines of code  & 1146 & 72 & 561 & 345 & 506 & 577 & 407 & 434 & 187 & 454 \\
Avg thinking time (s)& 1.6260 & 0.0000 & 0.0000 & 0.0440 & 0.7560 & 0.1780 & 0.0030 & 0.0040 & 1.7130 & 0.0050 \\
\hline
\multicolumn{11}{c}{\textbf{OIBench}} \\
Total input tokens & 1604531 & 40210 & 8364987 & 620122 & 665089 & 1081920 & 1027936 & 46854 & - & - \\
Complete tokens & 1610748 & 42112 & 8493081 & 653688 & 704924 & 1095605 & 1033256 & 78278 & - & - \\
Total output tokens & 6217 & 1902 & 128094 & 33566 & 39835 & 13685 & 5320 & 31424 & - & - \\
Total time (s) & 214.430 & 60.390 & 3604.330 & 3708.590 & 1235.390 & 411.730 & 259.790 & 810.060 & - & - \\
Tools used & 20 & 8 & 135 & 34 & 33 & 39 & 34 & 8 & - & - \\
Valid lines of code & 256 & 43 & 391 & 396 & 262 & 222 & 241 & 0 & 600 & 357 \\
Avg thinking time (s) & - & - & - & - & - & - & - & - & - & - \\
\hline
\end{tabular}
\end{sidewaystable}

%% file: tables/Gameprompt.tex
\begin{table}[b]
\centering
\caption{Game AI Development Prompt}
\label{tab:game_ai_prompt}
\begin{tabular}{p{0.95\linewidth}}
\toprule
\textbf{Game Development} \\
There is a game project under \textit{game env path}. You need to read its source code and develop a game AI. 
Your AI will compete against other AIs in a tournament, so please make your AI as strategic and competitive as possible. \\[6pt]

The final AI should be provided as an HTTP service. You can refer to the guides in 
\textit{game env path} / README.md and \textit{game env path} / develop\_instruction.md for development instructions. 
\textbf{The content in \textit{game env path} / develop\_instruction.md is very important, please read it carefully!} \\[6pt]
Please develop your AI service directly under \textit{dir path}.\\[6pt]
\textbf{Script Requirements} \\
Please implement a script to start your AI service, with the name \texttt{start\_ai.sh} in \textit{dir path}. The script must accept exactly one argument, which is the port number \textit{game port} to run the HTTP service. You should be able to start the AI service on a specified port by running: \\
\texttt{bash start\_ai.sh <port>} \\[6pt]
Your AI service should listen on the given port, and you can check its health status by running:
\texttt{curl -s http://localhost:<port>/health} \\[6pt]
\textbf{Note: } The script should not accept any other arguments except for the port number. Make sure your AI service uses this port for HTTP requests.\\[6pt]
\textbf{Other Requirements:} \\
Use your model name as a prefix in the name of your AI service, i.e.,  \textit{model name}\_AI. Develop directly in \textit{dir path} without repeatedly asking for the next step. Report to me only after you have completed the development. \\[6pt]

\textbf{Access the main server} \\
You can play the game at \textit{game server} using \textit{game env path}. You can play the games with your own AI or any other AI to improve your strategy. You can use bash tools to improve yourself. \\[6pt]

\textbf{Final Reminder} \\
You should write \texttt{start\_ai.sh} in \textit{dir path} and implement the AI service in \textit{dir path}. DO NOT MODIFY THE CODE IN \textit{game env path}\\

\textbf{Condition (if $round\_num > 1$):} \\
The tournament report from the last round is in \textit{last round log dir}. 
The historical records are quite large, please use tools \texttt{start\_interactive\_shell} and \texttt{run\_interactive\_shell} to analyze the data efficiently.  

The code corresponding to the log is stored in \textit{last round code dir}. 
Please learn from it and improve your strategy. \\
\bottomrule
\end{tabular}
\end{table}

%% file: tables/MLprompt.tex
\begin{longtable}{p{0.95\linewidth}}
\caption{Machine Learning Game AI with MANDATORY Self-Play Training Prompt}
\label{tab:ml_game_ai_prompt} \\
\toprule
\textbf{Machine Learning Game AI with MANDATORY Self-Play Training} \\
\midrule
\endfirsthead

\multicolumn{1}{c}{{\tablename\ \thetable{} -- continued from previous page}} \\
\toprule
\textbf{Machine Learning Game AI with MANDATORY Self-Play Training} \\
\midrule
\endhead

\midrule \multicolumn{1}{r}{{Continued on next page}} \\
\endfoot

\bottomrule
\endlastfoot

Develop a competitive game AI for \textit{game env path} using \textbf{REAL machine learning with actual training}. \\[6pt]

\textbf{CRITICAL: NO PSEUDO-ML ALLOWED} \\
\textbf{MANDATORY:} Implement real training with actual parameter updates. \\[6pt]

\textbf{Forbidden:} Random weights, unused optimizers, no training loops \\
\textbf{Required:} Self-play training, \texttt{loss.backward()}, \texttt{optimizer.step()}, saved trained model \\[6pt]

\textbf{Training Requirements} \\
1. \textbf{Self-Play System:} Generate training data by playing against itself \\
2. \textbf{Training Loop:} Real parameter updates with backpropagation \\
3. \textbf{Model Saving:} Save trained model weights (e.g., \texttt{trained\_model.pth}) \\
4. \textbf{Training Endpoint:} \texttt{/train} HTTP endpoint to trigger training \\[6pt]

\textbf{Technical Implementation} \\
The final AI should be provided as an HTTP service. You can refer to the guides in 
\textit{game env path}/README.md and \textit{game env path}/develop\_instruction.md for development instructions. 
\textbf{The content in \textit{game env path}/develop\_instruction.md is very important, please read it carefully!} \\
Please develop your AI service directly under \textit{dir path}. \\[6pt]

\textbf{Script Requirements} \\
Implement a script to start your AI service, with the name \texttt{start\_ai.sh} in \textit{dir path}. 
The script must accept exactly one argument, the port number to run the HTTP service. Start the AI via: \\
\texttt{bash start\_ai.sh <port>} \\[6pt]
Check service health via: \texttt{curl -s http://localhost:<port>/health} \\[6pt]

\textbf{Additional Training Script:} Also create \texttt{train\_ai.sh} for self-play training: \\
\texttt{bash train\_ai.sh <num\_episodes>} \\[6pt]

\textbf{Note:} The script should not accept any other arguments except for the port number. Ensure the AI service uses this port for HTTP requests. \\[6pt]

\textbf{Environment Requirements} \\
\textbf{MANDATORY:} You MUST develop under the \texttt{ml\_env\_py311} conda environment. 
PyTorch, TensorFlow, NumPy, pandas, and scikit-learn are already installed. \\
- DO NOT include PyTorch or TensorFlow in \texttt{requirements.txt} \\
- Create a requirements.txt file only for additional dependencies you need beyond the pre-installed packages \\
- All development and execution must be done in the \texttt{ml\_env\_py311} environment \\[6pt]

\textbf{Available Game Port} \\
We offer an interactive game port: \texttt{http://localhost:\textit{game port}} to facilitate your testing and training. You can interact with the service and get feedback from the port. \\[6pt]

\textbf{Validation Requirements} \\
Submission will be rejected if: \\
- The model uses only random weights without training \\
- No actual training loop with parameter updates \\
- No self-play data generation system \\
- Cannot demonstrate learning/improvement over time \\
- Training endpoints return fake responses without real training \\[6pt]

\textbf{Other Requirements} \\
Use your model name as a prefix in the name of your AI service, i.e., \textit{model name}\_AI. \\
Develop directly in \textit{dir path} without repeatedly asking for the next step. Report to me only after you have completed the development. \\[6pt]

\textbf{Final Reminder} \\
You should write \texttt{start\_ai.sh} and \texttt{train\_ai.sh} in \textit{dir path} and implement the REAL ML-based AI service with actual training capabilities in \textit{dir path}. DO NOT MODIFY THE CODE IN \textit{game env path}. 
Please make sure to implement a genuine machine learning training workflow, including writing the code, performing training, tuning hyperparameters, and testing the model. After training, save the trained model parameters, and ensure that the AI service can be started with the specified script to perform inference. \\[6pt]

\textbf{Mandatory:} Submission must include a trained model file (e.g., \texttt{trained\_model.pth}, \texttt{model\_weights.pkl}) proving real training. \\[6pt]

\textbf{Zero Tolerance for Pseudo-ML:} Any submission without real training will be rejected. \\
\end{longtable}

%% file: tables/Multilingualprompt.tex
\begin{table}[t]
\centering
\caption{Multilingual Game AI Development Prompt}
\label{tab:multila_game_ai_prompt}
\begin{tabular}{p{0.95\linewidth}}
\toprule
\textbf{Game Development} \\
There is a game project under \textit{game env path}. You need to read its source code and develop a game AI. 
Your AI will compete against other AIs in a tournament, so please make your AI as strategic and competitive as possible. \\[6pt]

The final AI should be provided as an HTTP service. You can refer to the guides in 
\textit{game env path} / README.md and \textit{game env path} / develop\_instruction.md for development instructions. 
\textbf{The content in \textit{game env path} / develop\_instruction.md is very important, please read it carefully!} \\[6pt]
Please develop your AI service directly under \textit{dir path}.\\[6pt]
\textbf{Script Requirements} \\
Please implement a script to start your AI service, with the name \texttt{start\_ai.sh} in \textit{dir path}. The script must accept exactly one argument, which is the port number \textit{game port} to run the HTTP service. You should be able to start the AI service on a specified port by running: \\
\texttt{bash start\_ai.sh <port>} \\[6pt]
Your AI service should listen on the given port, and you can check its health status by running:
\texttt{curl -s http://localhost:<port>/health} \\[6pt]
\textbf{Note: } The script should not accept any other arguments except for the port number. Make sure your AI service uses this port for HTTP requests.\\[6pt]
\textbf{Other Requirements:} \\
Use your model name as a prefix in the name of your AI service, i.e.,  \textit{model name}\_AI. Develop directly in \textit{dir path} without repeatedly asking for the next step. Report to me only after you have completed the development. \\[6pt]

\textbf{Access the main server} \\
You can play the game at \textit{game server} using \textit{game env path}. You can play the games with your own AI or any other AI to improve your strategy. You can use bash tools to improve yourself. \\[6pt]

\textbf{Final Reminder} \\
You should write \texttt{start\_ai.sh} in \textit{dir path} and implement the AI service in \textit{dir path}. DO NOT MODIFY THE CODE IN \textit{game env path}\\

\textbf{Condition (if language = JS)} \\
 JavaScript is the language you should use to develop your AI service. The version of Node.js is \textit{node version}, the path of Node.js is \textit{node path}, and it is already set in the PATH environment variable. 
You can use \texttt{node} to run the program. \\[6pt]

\textbf{Condition (if language = Go)} \\
 Go is the language you should use to develop your AI service. The version of Go is \textit{go version}, the path of Go is \textit{go path}, and it is already set in the PATH environment variable. 
You can use \texttt{go} to build the program. \\
\bottomrule
\end{tabular}
\end{table}